\documentclass[11pt]{article}

\usepackage[margin=1in]{geometry}
\usepackage{microtype}
\usepackage{booktabs}
\usepackage{enumitem}
\usepackage{macros}
\usepackage{algorithm}
\usepackage{algpseudocode}
\usepackage{xcolor}

\usepackage{graphicx}
\usepackage{pgfplots}
\usepackage{subcaption}
\pgfplotsset{compat=1.18}
\usepackage{tikz}
\usepackage{xcolor}
\definecolor{figblue}{HTML}{2764A5}
\definecolor{figpurple}{HTML}{7751B5}
\definecolor{figlavender}{HTML}{EEE8FA}
\newcommand{\Prb}{\mathbb P}
\newcommand{\one}{\mathbf 1}

\def\colorful{1}

\ifnum\colorful=1

\fi
\ifnum\colorful=0

\fi
\title{Geometrically Constrained Learning II:\\
Agnostic Proper Learning of Lipschitz Functions}

\author{
Jane Lange\thanks{Supported by the National Science Foundation
under Grant No. DMS-2424441} \\
Institute for Advanced Study \\
\texttt{janeclange@gmail.com}
\and
Cassandra Marcussen\thanks{School of Engineering and Applied Sciences, Harvard University, Cambridge, Massachusetts, USA. Email: cmarcussen@g.harvard.edu. This material is based upon work supported by the Air Force Office of Scientific Research under award number FA9550-23-F-0014 in the amount of \$139,400 to Cassandra Marcussen. Supported in part by NSF Award 2152413 and a
Simons Investigator Award to Madhu Sudan.} \\ Harvard University \\ \texttt{cmarcussen@g.harvard.edu}
\and
Arsen Vasilyan \\
Aarhus University \\
\texttt{vasilyan@cs.au.dk}
}
\date{}

\begin{document}
\maketitle
\begin{abstract}
We give a proper learner for 
Lipschitz functions over the 
Gaussian distribution on $\R^d$. Given random Gaussian examples labeled by a 
$\lambda$-Lipschitz function, $\lambda \ge 1$, the learner returns a function
$h:\R^d\to\R$ that
is globally $\lambda$-Lipschitz and has loss (absolute or squared) at most $\varepsilon$.
The sample complexity, running time, and representation
size are
\[
  (d+1)^{\poly(\lambda/\varepsilon)}.
\]
The learning algorithm can be made \emph{agnostic}, given a known
bound on the variance of the label marginal distribution. Some known
control of label scale is information-theoretically necessary.
For arbitrary labels with variance at most $V^2 \ge 1$, the complexity is
\[
  (d+1)^{\poly(V\lambda/\varepsilon)},
\]
and the error is $\mathsf{opt} + \varepsilon$, where $\mathsf{opt}$ is the minimal error 
among Lipschitz approximators. 

Our approach combines $L_1$ polynomial regression under a convex gradient constraint with local correction of the Lipschitz property.
The learner finds a polynomial that, after smoothing and clipping its 
range, is very close to a canonical Lipschitz correction of itself, the lower Lipschitz envelope.
Our local corrector modifies the lower envelope by replacing distances to sublevel sets with a Gaussian sampling proxy, 
allowing the construction to be locally and stably estimated from samples.
\end{abstract}

\newpage
\section{Introduction}
The Lipschitz property is a natural form of robustness for
real-valued functions: small perturbations to the input can cause
only proportionally small changes in the output. 
We study the task of learning Lipschitz functions over the Gaussian 
distribution on $\R^d$.
Formally, a $\lambda$-Lipschitz function $f:\R^d\to\R$ satisfies
\[
  |f(x)-f(y)|\le\lambda\|x-y\|_2
  \qquad\text{for all }x,y\in\R^d.
\]
In a machine learning context, this property enables several
practical safety and interpretability demands to be met.
Lipschitzness supplies the sensitivity bound needed for 
differentially private query
release \cite{JR2013}. 
A Lipschitz bound also limits the effect of perturbation-based backdoors; this is the principle
behind the smoothing defense of~\cite[Section~7]{GKVZ2022}, which trades
accuracy for a bound on sensitivity. 
Lipschitz constraints also play a role
in neural-network approaches to robustness and interpretable
gradients~\cite{ALG2019}.

Furthermore, because of the strength and simplicity of the property,
Lipschitzness also appears frequently as an assumption on the inputs
to other algorithmic and statistical problems.
Examples include learning and testing convex
functions~\cite{FPMMN2026}, monotonicity testing~\cite{FP2024}, metric-space bandits~\cite{KSU2008}, and global
optimization~\cite{MV2017}. 

\paragraph{Improper vs. proper learning.}
It is well known that Lipschitz functions are efficiently learnable by
polynomial regression, with polynomial dependence on dimension for fixed
accuracy and Lipschitz constant~\cite{FPMMN2026}.
However, predictive accuracy does not ensure that the learned 
hypothesis itself is Lipschitz, since prediction error is
measured on average under the input distribution, while the Lipschitz
property must hold for \textit{every} pair of inputs. A predictor can
agree closely with a Lipschitz function on almost all inputs and still
have arbitrarily large slopes on a small set of the domain. Thus, the violations may have little effect on the prediction error, but can crucially invalidate the stability or sensitivity assumption that a downstream
algorithm requires.

Such applications motivate \textit{proper learning}, where the returned
hypothesis must itself satisfy the Lipschitz condition. 
This is a computational problem, not a statistical one. Indeed, a net-based 
algorithm can restore the Lipschitz property after improper learning,
but it takes exponential time and outputs an exponential-sized description of its hypothesis
(cf.~\cite[Section~1.1.1]{FPMMN2026}). 
It is not clear a priori that a global Lipschitz constant can be enforced by a 
learning algorithm while preserving the $\poly(d)$ training and inference time
of polynomial regression. It is not even clear that every Lipschitz 
function can be approximated by one with a $\poly(d)$-sized description.
This leads to the question:
\begin{quote}
  \textit{Can we efficiently learn a Lipschitz function properly?}
\end{quote}
The backdoor-mitigation and privacy applications also motivate 
imposing Lipschitzness on a learned hypothesis
even when the labels do not themselves come from a Lipschitz function.
This yields the more general \emph{agnostic} learning question:
\begin{quote}
  \textit{Can we efficiently approximate arbitrary data by a Lipschitz hypothesis, competing with the best one?}
\end{quote}

\subsection{Our results}
We answer both questions affirmatively. We give an algorithm
that properly learns
$\lambda$-Lipschitz functions over the Gaussian distribution,
with $\poly(d)$ sample complexity, training time, and output
size. The theorems cited here are stated for $L_1$ error (absolute loss);
in \Cref{app:squared-loss} we show that these guarantees also hold for squared loss.

\begin{theorem}[Realizable proper learning (\Cref{thm:realizable}, informal)]
\label{thm:intro-realizable}
Let $g: \R^d\to\R$ be a $1$-Lipschitz function, $0<\eps\le1$, and $0<\delta<1/2$.
There is a randomized algorithm that takes random examples 
$(\bx, g(\bx)), \bx \sim \gauss$,
returning a function $H:\R^d\to\R$ such that
\[
 \Pr\!\left[
   \Lip(H)\le1
   \ \text{ and }\ 
   \err(H)\le\eps
 \right]\ge1-\delta.
\]
The sample complexity, running time, and output size are
\[
  (d+1)^{\poly(1/\eps)}\left(\log\frac2\delta\right)^3
\]
\end{theorem}
The algorithm extends to agnostic learning, given a bound on the
variance of the labels. 
Some known control of label scale is necessary for a uniform sample
bound; we discuss this in \Cref{app:scale-obstructions}. In the realizable setting, the needed
control follows directly from Lipschitzness.
Here $\OPT$ refers to the minimum error achieved by a Lipschitz function.
\begin{theorem}[Agnostic proper learning (\Cref{thm:unbounded-absolute}, informal)]
\label{thm:intro-agnostic}
Suppose $\operatorname{Var}(\by)\le V^2$ for a known
$V\ge1$, $0<\eps\le1$, and $0<\delta<1/2$.
There is a randomized algorithm
returning a finite formula $H:\R^d\to\R$ such that
\[
 \Pr\!\left[
   \Lip(H)\le1
   \ \text{ and }\ 
   \err(H)\le\OPT+\eps
 \right]\ge1-\delta.
\]
The sample complexity, running time, and output size are
\[
  (d+1)^{\poly(V/\eps)}\left(\log\frac2\delta\right)^3
\]
\end{theorem}
Scaling the labels and the output yields the result for general Lipschitz constants.\footnote{To obtain the corollary, apply the corresponding theorem to labels
$\by /\lambda$ with accuracy $\eps/\lambda$ for absolute loss and $\eps/\lambda^2$ for squared loss, and multiply the
returned function by $\lambda$. In the agnostic setting, use the
normalized variance bound $\max\{1,V/\lambda\}^2$.}
\begin{corollary}[General Lipschitz constants]
\label{cor:main-general-lipschitz}
For $\lambda\ge1$, the theorems above hold with the Lipschitz bound
$1$ replaced by $\lambda$ and, in the agnostic setting, with
\[
  \OPT_{\lambda}
  :=\inf_{\Lip(g)\le\lambda}\err(g)
\]
in place of $\OPT$. The complexity is
\[
  (d+1)^{\poly(\lambda/\eps)}\left(\log\frac2\delta\right)^3
\]
in the realizable setting, and
\[
  (d+1)^{\poly(V\lambda/\eps)}\left(\log\frac2\delta\right)^3
\]
in the agnostic setting.
\end{corollary}
To our knowledge, no subexponential-time proper learner was previously
known for the full class of Euclidean $1$-Lipschitz functions under
Gaussian inputs, even in
the realizable setting. Our algorithm achieves
polynomial dependence on the dimension for fixed accuracy, as in
polynomial-regression-based improper learning. All randomness is used
during training. The output is a deterministic, efficiently evaluable $\lambda$-Lipschitz function.

\subsection{Our techniques}\label{sec:proof-overview}
We briefly describe the primary algorithmic techniques; a more 
detailed walkthrough is given in \Cref{sec:proof-strategy}.
\paragraph{A convex constraint controlling gradients.}
Our algorithm performs polynomial regression with an 
additional constraint to control the distance to Lipschitzness.
We learn a polynomial $p$ that minimizes prediction error
subject to a constraint on a quantity we refer to as the 
squared \emph{gradient defect}:
\[
 \Gamma(p)\coloneqq\Ex_{\bx \sim \gauss}[(\|\nabla p(\bx)\|_2-1)_+^2].
\]
We prove that, when this defect is sufficiently small, a smoothed and
clipped version $q$ of $p$ is $1$-Lipschitz when restricted to
a set $S$ containing nearly all Gaussian mass. From this 1-Lipschitz partial function, it follows that the McShane extension\footnote{Distance to the empty set is $+\infty$.}
\[
 h(x)
 =\inf_{y\in S}\{q(y)+\|x-y\|_2\}
 =\inf_{a\in\R}
   \left\{a+\operatorname{dist}
     \bigl(x,\{y\in S:q(y)\le a\}\bigr)\right\},
\]
is a Lipschitz witness that agrees with $q$ on $S$.

\paragraph{Local correction.}
The above procedure provides a function $q$, a smoothed and clipped polynomial, that is close to Lipschitz in $L_0$ distance, 
but the witness (the McShane lower envelope) is not tractable to compute.
We do not compute this extension. Instead, we define an 
alternative ``lower envelope'' $\repair[q]$, such that $\repair[q](x)$ can be efficiently 
approximated at a specified $x$, given query access to $q$. This implements a 
\emph{local corrector} for the Lipschitz property.

The queries made by this corrector are samples from a small-scale 
Gaussian around $x$. The corrected value resembles the McShane
lower envelope, but the hard-to-compute distance term (Euclidean distance to the sublevel set of $a$) is replaced by an easy-to-approximate penalty term
derived from the Gaussian mass of the sublevel set.
\[
 \repair[q](x)
 =\inf_{a\in\R}
 \left\{
 a+\frac{\sigma}{\rho}
 \left[-\Phi^{-1}\!\left(
   \Prx_{\bz \sim \gauss} [q(\rho x+\sigma \bz)\le a]
 \right)\right]_+
 \right\},
 \qquad \rho=\sqrt{1-\sigma^2}.
\]
Here $t_+:=\max\{t,0\}$ denotes the positive part and $\Phi$ is the standard Gaussian CDF.
Thus each threshold offers its value plus a penalty for receiving
fewer than half the votes. We show that the corrected function is always
Lipschitz, and that functions that are close in $L_0$ distance
stay close in $L_1$ distance after correction.

We approximate the corrector using offsets drawn from a Gaussian
conditioned on a norm bound, once during training and then fixed; in other words, even
though the definition contains a random variable, the hypothesis
we output is deterministic. This output is not Lipschitz due to
instability in the approximation, but is close to Lipschitz in
$L_\infty$ distance. A final smoothing and rescaling step then
gives an exactly 1-Lipschitz output.

\subsection{A framework for enforcing notions of geometric robustness}
We view this proper Lipschitz objective as an instance of a 
broader class of adversarial robustness objectives that 
control the behavior of a learned hypothesis under a class of
perturbations. Typically, the notion of adversarial robustness 
is studied in classifiers, not regression problems, but
connections between Lipschitz regression and adversarially
robust classification are often brought up in the robust
learning literature \cite{CRK2019,GKVZ2022,SalmanLRZZBY19,ChiangEtAl2020}.

We apply the constrained regression techniques that we developed for 
this continuous prediction task to the robust classification problem
in a companion paper, \textit{Geometrically Constrained Learning I: Agnostically Learning Adversarially
Robust Classifiers}~\cite{LMV2026Robust}.
Both of these learning problems are about using the robustness of the
target class---its stability under OU smoothing, in the case of these
particular problems---to establish the existence of a low-degree
polynomial that is not only a good predictor, but restores a \textit{geometric
property} of the target class (in both cases, this property is 
represented as a gradient constraint). Stability under OU smoothing
furthermore allows the use of the Harnack inequality (see \Cref{eq:harnack}), a powerful
tool for transferring average-case bounds (on property violations, 
such as the gradient defect) to bounds that hold uniformly on large 
subsets of the domain. 

We believe that this toolkit may be more broadly applicable for learning
other types of robust target classes via geometric constraints, perhaps in other metric spaces
or with a different adversary model. 

\subsection{Related work}

\paragraph{Polynomial regression for agnostic learning.}
The $L_1$ polynomial regression framework of~\cite{KKMS2008}
converts low-degree approximation into improper agnostic learning;
\cite{KOS2008} relates this approximation to Gaussian surface area
for Boolean concept classes.

\paragraph{Learning Lipschitz functions.}
For general metric spaces, \cite{GKK2017} combines structural risk
minimization with approximate Lipschitz extension, with guarantees
that exploit low intrinsic dimension. Its sample and runtime bounds
depend exponentially on the doubling dimension
\cite[Remark~2]{GKK2017}. In Gaussian space,
\cite{FPMMN2026} learns a convex $1$-Lipschitz approximation to an
unknown $1$-Lipschitz function, achieving Gaussian $L_2$ distance
within $\eps$ of the optimum using $d^{O(1/\eps^2)}$ samples and
$\exp(\widetilde O(d/\eps^2))$ time. It first learns a Hermite
approximation, then enforces convexity and Lipschitzness.
We enforce Lipschitzness alone for the full Lipschitz class, with
both training and evaluation polynomial in $d$ for fixed accuracy.
Lipschitz-constrained GroupSort networks also give universal
approximation on compact domains~\cite{ALG2019}; this expressivity
result does not supply an efficient learning algorithm for the full
class.

\paragraph{Randomized smoothing for robust classification.}
Gaussian smoothing certifies robustness of classifiers to Euclidean
perturbations~\cite{CRK2019}. The inverse-Gaussian-CDF Lipschitz
bound of~\cite[Appendix~A, Lemma~2]{SalmanLRZZBY19} gives an
alternative proof of the result of \cite{CRK2019} and underlies our distance penalty. The work of~\cite{ChiangEtAl2020} introduces \emph{median smoothing} as a technique to improve the robustness of regression networks. 
Their paper applies the bound of \cite{SalmanLRZZBY19} to sublevel-set probabilities to bound the median-smoothed 
prediction under adversarial perturbations.

\paragraph{Local correction and local computation algorithms.}
Local correctors evaluate a function with a desired property using
few queries to an input function; they are a type of local computation algorithm~\cite{RTVX2011,ARVX2012}.  
The Lipschitz filters
of~\cite{JR2013} are local correctors; they preserve Lipschitz inputs on $\ell_1$ hypergrids
but have exponential query complexity in the dimension.
Subsequent work \cite{AJMR2014} proves exponential lower bounds for general
filters that always output a Lipschitz function. For bounded-range functions
on $\ell_1$ hypergrids, \cite{LLRV2025} give Lipschitz filters whose dimension exponents depend on the range
bound. 

Local correction of monotonicity, combined with constrained $L_1$ polynomial regression,
yields agnostic proper learning of monotone Boolean functions
under the uniform distribution~\cite{LRV2022,LV2023}; this line
of work introduces the local-correction strategy for proper learning.

\subsection{AI methodology}

We used OpenAI's GPT-6 Astra, GPT-6.1 Sol, and GPT-5.6 Sol to assist with 
developing and checking arguments, drafting and revising the 
exposition, and searching for related work.
This use materially affected the technical sections of the paper
and the list of references. The authors take full responsibility for all results,
proofs, references, and claims of originality.

\section{Technical overview}\label{sec:proof-strategy}
The algorithm's high-level structure is to first learn a polynomial,
then use local correction to enforce Lipschitzness.
The first step outputs a polynomial with small prediction
loss and small \textit{gradient defect}, which measures how much
the norm of the gradient exceeds 1 on average (see \Cref{eq:gradient-excess}). 
Gradient defect is convex in the coefficients of the polynomial,
so standard convex optimization techniques allow us to obtain these 
two guarantees simultaneously.

The main challenge comes after this step. Even if we are given a
polynomial with extremely small defect, it is
not clear how to turn it into an efficiently evaluable Lipschitz
function. Its gradient norm may exceed 1 on a tiny region that is
unlikely to be seen in samples. A single such region can violate
the global Lipschitz condition.

We address this challenge using two stages of local correction.
The first stage computes a type of lower envelope that can be efficiently
estimated with Gaussian samples. If we had access to exact population 
statistics, then this envelope would already be Lipschitz, and the 
second corrector would be unnecessary. When implemented using a finite
sample, we instead get a pointwise bound on distance to Lipschitzness.
The second corrector smooths this output to transfer the pointwise
bound to a global Lipschitz constant slightly greater than 1, then
rescales to restore 1-Lipschitzness.

For notational convenience, we present our algorithm as if it were 
learning a \emph{bounded} 1-Lipschitz function $f: \R^d \to [-B,B]$.
For correction, we work with $p-\E p$, writing this centered
polynomial as $p$ in the overview, and restore its center afterward.
Label centering is handled separately.
Centering and clipping the labels can be handled separately, and this
reduces unbounded realizable labels to bounded
realizable labels. This argument is given in \Cref{sec:main-proof}. 
The agnostic and squared-loss extensions are given in \Cref{app:agnostic,app:squared-loss}.
%

\subsection{Learning a polynomial with controlled loss and gradient defect}
We find a polynomial $p$ with small prediction loss and
small squared gradient defect:
\begin{equation}\label{eq:gradient-excess}
  \E|p(\bx)-\by|\le\eta,
  \qquad
  \Gamma(p):=
  \E\bigl[(\|\nabla p(\bx)\|_2-1)_+^2\bigr]
  \le\alpha^2.
\end{equation}
For a differentiable function, the Lipschitz condition
is equivalent to requiring that its gradient norm be at most 1
everywhere. The quantity $\Gamma(p)$ is a relaxation of this
condition that we can optimize using samples. In Step~2, we show
that making it sufficiently small ensures proximity to a
Lipschitz function.

Our polynomial learner minimizes prediction loss subject to a bound
on (an empirical estimate of) $\Gamma(p)$. Both the objective and the
constraint are convex in the polynomial coefficients. 
To show that this objective has a good solution, we must argue that
every bounded Lipschitz function is close to a low-degree polynomial with small
defect. Let $f$ be a bounded Lipschitz function; we obtain this polynomial $Q = (U_t f)^{\le k}$
witness by applying Ornstein--Uhlenbeck smoothing at a small scale $t$,
then truncating Hermite expansion to degree $k$.

Lipschitz functions are noise-stable and OU smoothing preserves
Lipschitzness, so $f$'s Lipschitz bound ensures
that smoothing changes its predictions only slightly and keeps its 
gradient norm at most 1. Thus, $Q$'s gradient norm can exceed one 
only because of the truncation error. 

OU smoothing causes degree-$k$ Hermite coefficients (of $Q$ and its derivatives)
to decay exponentially, as $e^{-tk}$. Thus, truncating at degree
$O(t^{-1}\log(B/(\alpha\eta)))$ introduces
$\approx \eta \cdot \alpha$ error to the predictions
and the gradient. Because of the logarithmic degree dependence,
we can obtain
$\Gamma(p)\le\alpha^2$ with $\alpha=\exp(-\Omega(1/\eps^4))$, while keeping the degree polynomial in
the label bound and inverse accuracy.

\subsection{Establishing $L_0$ closeness to a Lipschitz function}
We smooth the learned polynomial $p$ once more and then clip its
values to the label range. Write
\[
  P=U_sp,
  \qquad
  q=\cB(P),
\]
where $U_s$ is the Ornstein--Uhlenbeck operator with a new smoothing
time $s$, and $\cB$ clips
to $[-B,B]$. These operations change the prediction loss only
slightly. Small gradient defect ensures that smoothing changes
$p$ by only $O(\sqrt{s})$ in Gaussian $L_2$. Clipping cannot
increase loss, so the additional prediction loss is at
most $O(B\sqrt{s})$.

We next show that $q$ agrees with a Lipschitz function on most
inputs. Write $r=(\|\nabla p\|_2-1)_+$. Smoothing contracts
gradients, giving
$\|\nabla P(z)\|_2\le e^{-s}(1+U_sr(z))$.
The Gaussian Harnack inequality
\cite[Section~5.6]{BakryGentilLedoux2014} gives
\begin{equation}\label{eq:harnack}
    U_sr(z)\le \sqrt{U_s(r^2)(x)}
  \exp\!\left(\frac{\|z-x\|_2^2}{2(e^{2s}-1)}\right).
\end{equation}
If $U_s(r^2)(x)$ is sufficiently small, this bounds $U_sr(z)$
by $e^s-1$ throughout the radius-$2B$ ball around $x$.
Thus $P$, and therefore $q$, is $1$-Lipschitz on this ball.
For more distant points, boundedness gives
$q(x)-q(y)\le2B\le\|x-y\|_2$. Put together, at every such
good center $x$,
\[
  q(x)\le q(y)+\|x-y\|_2
  \qquad\text{for every }y\in\mathbb{R}^d.
\]

This is exactly the condition under which $q$ agrees with its
\textit{McShane envelope} \cite{McShane1934}:
\[
  h(x):=\inf_{y\in\mathbb{R}^d}
  \{q(y)+\|x-y\|_2\}.
\]
For each fixed $y$, the expression inside the infimum is
$1$-Lipschitz as a function of $x$, and taking the infimum
preserves this bound. Also, $h\le q$, by choosing $y=x$.
At a good center, the inequality above gives the reverse bound,
so $h(x)=q(x)$. The envelope is therefore a globally
$1$-Lipschitz function that agrees with $q$ on a large set $S$.
It also takes values in $[-B,B]$.

Markov's inequality and Gaussian stationarity control the measure
of the exceptional set $S^c$:
\[\mu:=\Pr[q(\mathbf x)\ne h(\mathbf x)]
\le C_{s,B}\Gamma(p),\]
where $C_{s,B}$ grows exponentially in $B^2/s$, due to exponential dependence on $\norm{z-x}^2/s$ in the Harnack inequality.
We choose the gradient defect small enough that $\mu$ remains small
after both this amplification and the further amplification in the next step.

The definition of the envelope $h$ is intractable to compute; it
requires a global minimization over $\mathbb{R}^d$. It is only an 
analysis tool and the algorithm never computes it. Our next step is
to construct a local correction that stays close to $h$ and
admits an efficient implementation.

\subsection{Constructing a local Gaussian correction}

We now correct $p$ locally to obtain a function uniformly close
to a single Lipschitz function.

We assumed that $p$ is centered at 0 so, Gaussian Poincar\'e gives
$\|p\|_2\le1+\sqrt{\Gamma(p)}=O(1)$.
For correction, choose an internal clipping height $L=\varepsilon^{-1}$,
separate from the label bound $B$. Apply the step of the prior section with $L$ in place of $B$,
retaining $P=U_sp$ and redefining $q$ by
\[
 q=C_L(P).
\]
The previous step gives a $1$-Lipschitz McShane witness
$h:\mathbb R^d\to[-L,L]$ that agrees with $q$ outside a small set.

Use the same smoothing time $s$ for the probes
$\bw_x=\rho_s x+\sigma_s \bz$, where $\bz\sim\gauss$,
$\rho_s=e^{-s}$, and $\sigma_s=\sqrt{1-e^{-2s}}$.
For $v:\mathbb R^d\to[-L,L]$, define
\[
 \begin{aligned}
		 \text{vote probability: } \quad &q_a[v](x) =\mathbb P[v(\bw_x)\le a],\\
		 \text{penalty:}\quad &\psi(r) =\min\left\{2L,\frac{\sigma_s}{\rho_s}
                       [-\Phi^{-1}(r)]_+\right\},\\
				 \text{corrected value: }\quad &\repair[v](x) =\inf_{a\in[-L,L]}
                    \{a+\psi(q_a[v](x))\},
 \end{aligned}
\]
with $\psi(0)=2L$ and $\psi(1)=0$.
Each threshold pays its value plus a penalty for receiving fewer
than half the votes. The cap of $2L$ makes $\psi$ Lipschitz, with Lipschitz constant
$K=\exp(O(L^2/s))$ for small $s$, without changing the infimum.

We call $\repair[v]$ the \emph{ideal correction} of $v$.
Like the McShane extension, it minimizes a threshold plus a cost
associated with its sublevel set. Here the cost comes from a
Gaussian hitting probability, which we can estimate by sampling.
\begin{figure}[H]
\centering
\captionsetup{font=footnotesize}
\captionsetup[subfigure]{font=footnotesize}

\begin{subfigure}[t]{0.47\linewidth}
\centering
\begin{tikzpicture}[font=\scriptsize]
\begin{axis}[
    width=\linewidth,
    height=0.75\linewidth,
    axis lines=left,
    xmin=0, xmax=1,
    ymin=-0.08, ymax=2.3,
    xlabel={Vote probability $r$},
    ylabel={Penalty $\psi(r)$},
    xtick={0,0.5,1},
    xticklabels={$0$,$1/2$,$1$},
    ytick={0,1,2},
    yticklabels={$0$,$1$,$2L$},
    font=\scriptsize,
    label style={font=\scriptsize},
    tick label style={font=\scriptsize},
]
\addplot[black, thick]
    coordinates {(0,2) (0.0227501,2)};

\addplot[black, thick, smooth] coordinates {
    (0.0227501,2.0)
    (0.0359303,1.8)
    (0.0547993,1.6)
    (0.0807567,1.4)
    (0.1150697,1.2)
    (0.1586553,1.0)
    (0.2118554,0.8)
    (0.2742531,0.6)
    (0.3445783,0.4)
    (0.4207403,0.2)
    (0.5000000,0.0)
};

\addplot[black, thick]
    coordinates {(0.5,0) (1,0)};

\draw[gray, dotted]
    (axis cs:0.5,0) -- (axis cs:0.5,2.15);

\node at (axis cs:0.17,2.13) {cap};
\draw[gray]
    (axis cs:0.10,2.08) -- (axis cs:0.012,2);

\node at (axis cs:0.75,0.25) {no penalty};
\end{axis}
\end{tikzpicture}
\caption{The penalty for $L=c_s=1$.}
\label{fig:gaussian-penalty}
\end{subfigure}%
\hfill%
\begin{subfigure}[t]{0.47\linewidth}
\centering
\begin{tikzpicture}[font=\scriptsize]
\begin{axis}[
    width=\linewidth,
    height=0.75\linewidth,
    axis lines=left,
    xmin=-1, xmax=1,
    ymin=-1.08, ymax=1.25,
    xlabel={Threshold $a$},
    ylabel={Value},
    xtick={-1,0,0.25,1},
    xticklabels={$-1$,$0$,$m$,$1$},
    ytick={-1,0,1},
    font=\scriptsize,
    label style={font=\scriptsize},
    tick label style={font=\scriptsize},
    legend style={
        draw=none,
        font=\tiny,
        at={(0.98,0.03)},
        anchor=south east,
    },
]
\addplot[gray, dashed, thick, domain=-1:1] {x};
\addlegendentry{Threshold $a$}

\addplot[black, thick] coordinates {
    (-1,0.76776695)
    (0.25,0.25)
    (1,1)
};
\addlegendentry{Score}

\draw[gray, dotted]
    (axis cs:0.25,-1.08) -- (axis cs:0.25,0.25);
\fill[black] (axis cs:0.25,0.25) circle (1.5pt);

\node[anchor=south]
    (median) at (axis cs:-0.12,1.02)
    {minimum $=m$};
\draw[->, darkgray]
    (median.south) -- (axis cs:0.25,0.29);

\draw[<->, darkgray]
    (axis cs:-0.6,-0.6) -- (axis cs:-0.6,0.6020815);
\node[anchor=west]
    at (axis cs:-0.55,-0.18) {penalty};
\end{axis}
\end{tikzpicture}
\caption{The minimum score equals the median.}
\label{fig:gaussian-score}
\end{subfigure}

\caption{
The ideal Gaussian corrector.
(a) A threshold with vote probability $r\ge1/2$ pays no
penalty; below $1/2$, the penalty is
$\min\{2L,-c_s\Phi^{-1}(r)\}$.
(b) Threshold scores for
$v(y)=\max\{-1,\min\{1,y\}\}$, with
$\rho_s=\sigma_s=1/\sqrt{2}$ and $\rho_s x=1/4$.
The dashed line is the threshold value $a$; the solid line
is its penalized score. The minimum occurs at the Gaussian
median $m=1/4$: lowering the threshold below $m$ costs more
in penalty than it saves in threshold value.
}
\label{fig:gaussian-corrector}
\end{figure}
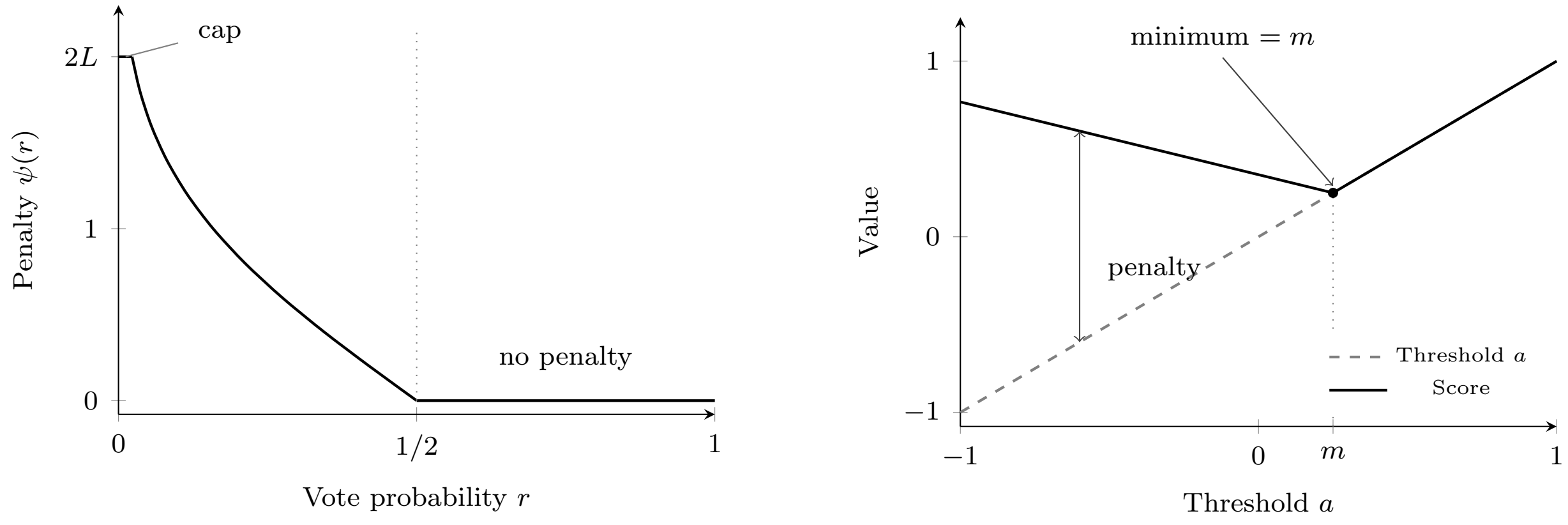
\paragraph{3(a). The ideal correction is $1$-Lipschitz.}
Each vote function $q_a[v]$ is a Gaussian average of
$\mathbf1_{\{v\le a\}}$. For a nonconstant vote function,
the Gaussian smoothing inequality \cite{BakryGentilLedoux2015} gives
\[
 |\Phi^{-1}(q_a[v](x))-\Phi^{-1}(q_a[v](y))|
 \le\frac{\rho_s}{\sigma_s}\|x-y\|.
\]
Multiplying by $\sigma_s/\rho_s$, taking the negative part, and capping
make each threshold score $1$-Lipschitz. Constant vote functions
give constant scores. Taking the infimum preserves the bound.

\paragraph{3(b). The ideal correction preserves accuracy.}
We compare $q$ to its repair through the Lipschitz witness $h$, which is the McShane extension of $q$ given in Step 2.

First, $\repair[h]$ equals the lower median $m(x)$ of $h(\bw_x)$.
The threshold $m(x)$ pays no penalty, so its score is $m(x)$.
Higher thresholds cannot improve it. For $a<m(x)$, Gaussian isoperimetry \cite{Borell1975} gives
$\psi(q_a[h](x))\ge m(x)-a$, so lower thresholds cannot improve
the score either. Gaussian noise stability \cite[Proposition~11.37]{ODonnell2021}
bounds the mean-square change under perturbation:
\[
 \E_{\bx,\bz}|h(\bw_{\bx})-h(\bx)|^2\le2(1-\rho_s).
\]
A median can be far from $h(x)$ only if at least half the probe
values are at least as far away. Thus
\[
 \|\repair[h]-h\|_2^2
 =\|m-h\|_2^2
 \le2\E_{\bx,\bz}|h(\bw_{\bx})-h(\bx)|^2
 \le4(1-\rho_s).
\]

Next, let $E=\{q\ne h\}$ and $\mu=\mathbb P[\bx\in E]$.
Only probes landing in $E$ can change a vote, so
\[
 |q_a[q](x)-q_a[h](x)|\le\mathbb P[\bw_x\in E].
\]
Recall that the penalty function is $K$-Lipschitz,
where $K=\exp(O(L^2/s))$.
Applying the $K$-Lipschitz penalty and taking infima changes this
bound by at most a factor of $K$. Since $\rho_s \bx+\sigma_s \bz$ is
standard Gaussian, integrating gives
\[
 \|\repair[q]-\repair[h]\|_1\le K\mu.
\]
Also, $\|h-q\|_1\le2L\mu$. Combining the three comparisons,
\[
 \|\repair[q]-q\|_1
 \le (K+2L)\mu+2\sqrt{1-\rho_s}.
\]
With $L=\varepsilon^{-1}$ and $s=2\varepsilon^2$, we have
$\varepsilon\le\sigma_s\le2\varepsilon$ and $K=\exp(O(\varepsilon^{-4}))$.
This is the second place where the exponentially small gradient defect is used:
our choice of $\alpha$ ensures $\mu\le\exp(-C/\varepsilon^4)$ for a sufficiently
large $C$, so $(K+2L)\mu=O(\varepsilon)$. The median, smoothing, and clipping
errors are also $O(\varepsilon)$, giving
\[
 \|\repair[q]-p\|_1=O(\varepsilon).
\]

\paragraph{3(c). Approximate the correction by sampling.}
Fix a desired pointwise accuracy $0<\beta\le\varepsilon$.
Store independent Gaussian vectors $\bz_1,\ldots,\bz_N$
conditioned on a deterministic norm bound.
At query $x$, sort the values $u_i=q(\rho_s x+\sigma_s \bz_i)$ and return
\[
 G(x)=\widehat{\repair}[q](x),\qquad
 \widehat{\repair}[q](x)
 =\min_{j\le N}\{u_{(j)}+\psi(j/N)\}.
\]
This replaces ideal vote probabilities by empirical frequencies.
For a fixed query $x$ and threshold $-L\le a<L$, a probe $z$
votes for $a$ exactly when
\[
 P(\rho_s x+\sigma_s z)\le a.
\]
This is a polynomial inequality of degree at most $k$ in $z$.
If we represent $z$ by the vector of all its monomials of degree
at most $k$, the inequality becomes linear. Thus each vote event
is a halfspace in this feature space. At $a=L$, every probe votes. A VC argument shows
that sufficiently many samples approximate every vote probability
within $\beta/K$, simultaneously over all queries and thresholds,
with high probability. The $K$-Lipschitz penalty then gives
\[
 \sup_x|\widehat{\repair}[q](x)-\repair[q](x)|\le\beta.
\]
Thus $G$ is $O(\varepsilon)$-close to $p$ in $L_1$ and uniformly
$\beta$-close to the single $1$-Lipschitz function $\repair[q]$.

\subsection{Obtaining an exactly Lipschitz output}

We still need to control slopes in order to output a Lipschitz
function. The output of the empirically-approximated corrector is
uniformly $\beta$-close to Lipschitz, but its slopes may still be arbitrarily large. Thus, we perform one more step of OU smoothing.

Think of $G$, the corrector output, as the sum of the nearby Lipschitz 
function $\repair[q]$ and a residual function $r:\R^d \to [-\beta,\beta]$. 
Smoothing $r$ at time $\beta$ reduces its gradient to $\sqrt{\beta}$,
which results in a $(1+\sqrt{\beta})$ Lipschitz constant. Rescaling by a
factor of $1/(1+\sqrt{\beta})$ would then complete the correction.

This would be sufficient, except it is not efficiently computable. Thus we replace the Gaussian
expectation by a finite average $A$, using the same probes for
every query. We must approximate gradients as well as values in comparison with the smoothed version of $G$,
because value approximation alone would leave the slopes of $A$
uncontrolled.

The piecewise polynomial form of $G$ enables us to approximate
both values and gradients using empirical estimates.
Polynomial growth bounds control the sampled gradients, while
the polynomial comparisons used to select a piece give
pseudo-dimension bounds. Uniform convergence then ensures that
the empirical averages approximate both values and gradients
simultaneously for every query in $\mathcal B_R$.

We prove that, with high probability, $A$ is $(1+2\sqrt\beta)$-Lipschitz on a ball
$\mathcal B_R$ containing nearly all Gaussian mass.
Projecting onto this ball and rescaling gives
\[
 G_{\mathrm{prop}}(x)
 =\frac{A(\Pi_Rx)}{1+2\sqrt\beta},
 \qquad
 \|G_{\mathrm{prop}}-G\|_2\le8(L+1)\sqrt\beta.
\]
The output is exactly $1$-Lipschitz, and choosing $\beta$ small
enough completes the error bound.

\begin{figure}[h]
\centering
\begin{tikzpicture}[x=1.05cm,y=0.84cm]
  \begin{scope}
    \node[font=\small,anchor=west] at (-2.55,1.62)
      {(a) Pointwise closeness};
    \draw[gray!55,thin,->] (-2.65,0) -- (2.7,0)
      node[below left,font=\scriptsize] {$x$};
    \draw[gray!55,thin,->] (-2.55,-1.26) -- (-2.55,1.34);
    \fill[figlavender]
      plot[domain=-2.5:2.5,samples=90]
        (\x,{0.34*\x+0.15*sin(deg(0.9*\x))+0.19})
      -- plot[domain=2.5:-2.5,samples=90]
        (\x,{0.34*\x+0.15*sin(deg(0.9*\x))-0.19}) -- cycle;
    \draw[figpurple,thick] plot[domain=-2.5:2.5,samples=240]
      (\x,{0.34*\x+0.15*sin(deg(0.9*\x))
          +0.13*sin(deg(10*\x))+0.035*sin(deg(21*\x))});
    \draw[figblue,dashed,thick] plot[domain=-2.5:2.5,samples=90]
      (\x,{0.34*\x+0.15*sin(deg(0.9*\x))});
    \node[font=\scriptsize,anchor=west,fill=white,inner sep=1pt]
      at (-2.38,-1.58)
      {{\color{figpurple}solid: $G$}
       \quad {\color{figblue}dashed: $h_\star$}};
  \end{scope}
  \begin{scope}[xshift=6.5cm]
    \node[font=\small,anchor=west] at (-2.55,1.62)
      {(b) Smooth and rescale};
    \draw[gray!55,thin,->] (-2.65,0) -- (2.7,0)
      node[below left,font=\scriptsize] {$x$};
    \draw[gray!55,thin,->] (-2.55,-1.26) -- (-2.55,1.34);
    \draw[figpurple,thick,dashed] plot[domain=-2.5:2.5,samples=90]
      (\x,{0.332*\x+0.137*sin(deg(0.9*\x))});
    \draw[figblue,thick] plot[domain=-2.5:2.5,samples=90]
      (\x,{0.290*\x+0.120*sin(deg(0.9*\x))});
    \node[font=\scriptsize,anchor=west,fill=white,inner sep=1pt]
      at (-2.38,-1.58)
      {{\color{figpurple}dashed: $A$}
       \quad {\color{figblue}solid: $G_{\mathrm{prop}}$}};
  \end{scope}
\end{tikzpicture}
\caption{The correction, schematically. Although $G$ stays within a
small band around the specified Lipschitz witness $h_\star$, it can have steep local oscillations.
Gaussian smoothing gives a nearly $1$-Lipschitz function.
Approximating its values and gradients by a finite average $A$,
then projecting and rescaling, gives $G_{\mathrm{prop}}$.}
\label{fig:section7-smoothing}
\end{figure}
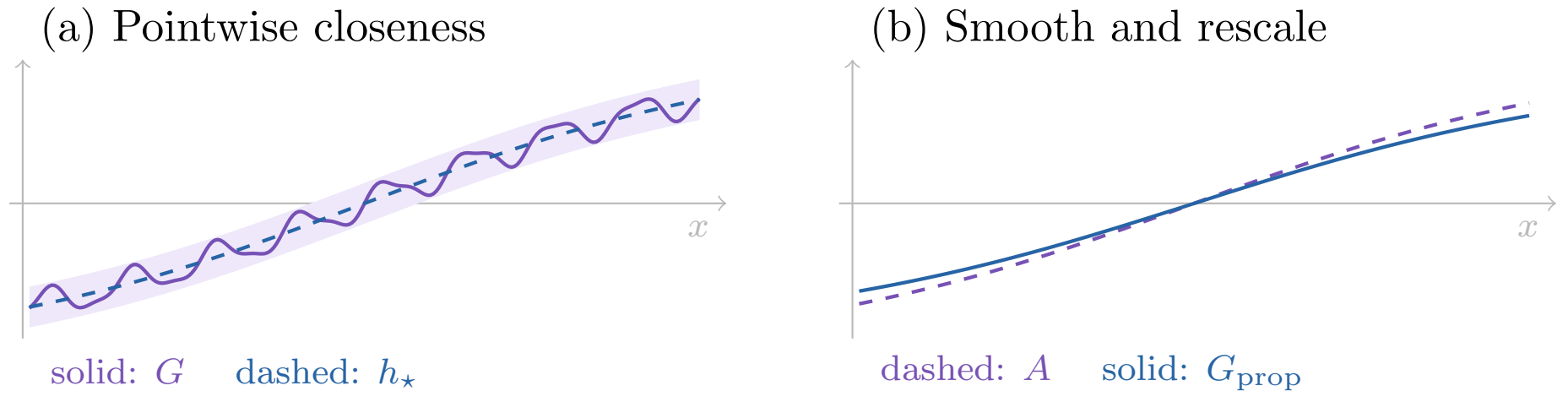

\section{Preliminaries}
\label{sec:preliminaries}

Throughout the main body, we consider absolute loss in the realizable
setting: $\bx\sim\gauss$ and $\by=g(\bx)$ for a $1$-Lipschitz function
$g:\R^d\to\R$. Thus $\err_1(f)=\E|f(\bx)-\by|$ and $\OPT_1=0$.
The polynomial learner first handles the bounded-label subproblem
$g:\R^d\to[-B,B]$, with known $B\ge1$.
\Cref{sec:realizable-clipping} reduces the general case to this subproblem,
and \Cref{thm:realizable} gives the final guarantee without a label bound.
Agnostic learning and squared loss are treated in
\Cref{app:agnostic,app:squared-loss}.
The $L_2$ norms below are used to control approximation error and gradients.

\subsection{Notation and Gaussian facts}

We use bold lowercase symbols (e.g. $\bx$, $\by$, and $\bz$) for
random variables, and ordinary symbols for fixed inputs and parameters.
All unqualified function norms and inner products are with respect to
$\gauss$, and vector norms are Euclidean. We write
\[
  \cB(t)=\max\{-B,\min\{B,t\}\}
\]
for clipping to $[-B,B]$, and
\[
  \Gamma(p)
  =
  \E\pos{\norm{\nabla p(\bx)}-1}^{2}
\]
for the squared gradient defect of $p$. Thus $\Gamma(p)$ measures
how much the gradient norm exceeds the target Lipschitz constant~$1$.
We denote the class of bounded $1$-Lipschitz functions by
\[
  \calL_B
  =
  \{h:\R^d\to[-B,B]:\Lip(h)\le1\}.
\]
For $r>0$, $\proj_r$ denotes Euclidean projection onto the closed ball
of radius $r$ centered at the origin.

\paragraph{Hermite expansions.}
Let $H_\nu$ be the normalized Hermite polynomial indexed by
$\nu\in\mathbb Z_{\ge0}^d$, with total degree
$|\nu|=\sum_{i=1}^d\nu_i$. These polynomials form an orthonormal
basis of $L_2(\gauss)$. For $f\in L_2(\gauss)$, write
\[
  \widehat f(\nu)=\langle f,H_\nu\rangle,
  \qquad
  f_j=\sum_{|\nu|=j}\widehat f(\nu)H_\nu,
  \qquad
  f=\sum_{j\ge0}f_j,
  \qquad
  f^{\le k}=\sum_{j\le k}f_j.
\]
Here $\widehat f(\nu)$ is a scalar coefficient, whereas $f_j$ is the
entire degree-$j$ component. Orthonormality and differentiation give
\[
  \norm f_2^2=\sum_\nu\widehat f(\nu)^2,
  \qquad
  \partial_iH_\nu=\sqrt{\nu_i}\,H_{\nu-e_i},
\]
where $e_i$ is the $i$th standard basis vector and the derivative is
zero when $\nu_i=0$. See \cite[Chapter~11]{ODonnell2021} for background.

\paragraph{Ornstein--Uhlenbeck smoothing.}
For $s\ge0$, write $\rho_s=e^{-s}$ and $\sigma_s=\sqrt{1-e^{-2s}}$.
The Ornstein--Uhlenbeck operator averages a function under a Gaussian
perturbation of its input:
\[
  U_sf(x)
  =
  \E_{\bz}  f\!\left(
    \rho_s x+\sigma_s \bz
  \right),
  \qquad \bz\sim\gauss.
\]
It preserves the Gaussian measure and contracts $L_2(\gauss)$.
In the Hermite basis,
\begin{equation}\label{eq:ou-identities}
  U_sf=\sum_{j\ge0}e^{-sj}f_j,
  \qquad
  \norm{\nabla f}_2^2
  =
  \sum_{j\ge1}j\norm{f_j}_2^2,
  \qquad
  \nabla U_sf=\rho_s U_s(\nabla f).
\end{equation}
The gradient identities hold for functions in $L_2(\gauss)$ whose
weak gradients also lie in $L_2(\gauss)$; in particular, they apply
to polynomials and Lipschitz functions
\cite[Chapter~11]{ODonnell2021}. If $f$ is $1$-Lipschitz, then
$\norm{\nabla f}_2\le1$.

We also use the correlation parametrization
\[
  T_\rho f(x)
  =
  \E_{\bz}  f\!\left(\rho x+\sqrt{1-\rho^2}\,\bz\right),
  \qquad 0\le\rho<1.
\]
Thus $T_\rho=U_s$ when $\rho=\rho_s$. We write $\by\sim_\rho x$ when
\[
  \by=\rho x+\sqrt{1-\rho^2}\,\bz,
  \qquad \bz\sim\gauss,
\]
and call $\by$ a noisy sample from $x$ at correlation $\rho$.

\paragraph{Polynomial growth.}
We will need a crude bound on polynomial values and gradients
at individual points.

\begin{fact}[Polynomial growth]
\label{fact:polynomial-growth}
For every degree-at-most-$k$ polynomial $P$ and every $y\in\R^d$,
\[
  |P(y)|+\norm{\nabla P(y)}
  \le
  \norm P_2[C(d+k+1)]^{k+1}(1+\norm y)^k,
\]
where $C$ is universal. Also, for $\bz\sim\gauss$,
\[
  \E(1+\norm \bz)^{2k}\le[C(d+k+1)]^k.
\]
\end{fact}

\begin{proof}
The normalized univariate Hermite polynomial has the formula
\[
  H_j(u)
  =
  \sqrt{j!}\sum_{m=0}^{\lfloor j/2\rfloor}
  \frac{(-1)^m u^{j-2m}}{2^m m!(j-2m)!}.
\]
Taking products over coordinates gives
\[
  |H_\nu(y)|
  \le
  [C\sqrt{k+1}(1+\norm y)]^{|\nu|}
  \qquad (|\nu|\le k).
\]
Expand $P$ in the $\binom{d+k}{k}$ Hermite polynomials of degree
at most $k$. Cauchy--Schwarz, orthonormality, and the derivative
identity above give the first bound, after enlarging $C$.
The second follows from
$(\E\norm \bz^{2k})^{1/(2k)}\le C\sqrt{d+k}$ for $k\ge1$;
the case $k=0$ is immediate.
\end{proof}

\paragraph{Gaussian isoperimetry.}
Let $\varphi$ and $\Phi$ denote the standard one-dimensional Gaussian
density and CDF, respectively, and let
\[
  I(r)=\varphi(\Phi^{-1}(r))
\]
be the Gaussian isoperimetric function. We use the following
standard enlargement inequality.

\begin{fact}[Gaussian enlargement inequality \cite{Borell1975}]
\label{fact:isoperimetric}
Let $A\subseteq\R^d$ be measurable and, for $r>0$, let
\[
  A_r=\{x\in\R^d:\dist(x,A)\le r\}.
\]
Then
\[
  \gauss(A_r)
  \ge
  \Phi\!\left(\Phi^{-1}(\gauss(A))+r\right).
\]
\end{fact}

\subsection{Facts for transferring risk}
\label{sec:risk-transfer}

We use the following facts to translate approximation bounds into
risk bounds.

\begin{fact}[Clipping and risk transfer]
\label{fact:risk-transfer}
Suppose $|\by|\le B$. Clipping a predictor to $[-B,B]$ cannot increase
absolute loss. For any integrable predictors $f,g$,
\[
  |\err_1(f)-\err_1(g)|
  \le\norm{f-g}_1
  \le\norm{f-g}_2.
\]
\end{fact}

\begin{proof}
Clipping decreases the distance to every $y\in[-B,B]$.
The risk bound follows by integrating
$\bigl||f-\by|-|g-\by|\bigr|\le|f-g|$ and applying Cauchy--Schwarz.
\end{proof}

\begin{fact}[OU smoothing cost]
\label{fact:ou-smoothing-cost}
For every $f\in L_2(\gauss)$ with weak gradient in $L_2(\gauss)$
and every $s\ge0$,
\begin{equation}\label{eq:ou-smoothing-cost}
  \norm{f-U_sf}_2
  \le
  \sqrt{1-e^{-s}}\,\norm{\nabla f}_2
  \le
  \sqrt{s}\,\norm{\nabla f}_2.
\end{equation}
In particular, $\norm{h-U_sh}_2\le\sqrt{s}$ for every
$1$-Lipschitz function $h$.
\end{fact}

\begin{proof}
The OU identities give
\[
  \norm{f-U_sf}_2^2
  =
  \sum_{j\ge1}(1-e^{-sj})^2\norm{f_j}_2^2.
\]
Since
$(1-e^{-sj})^2\le1-e^{-sj}\le j(1-e^{-s})$,
\[
  \norm{f-U_sf}_2^2
  \le
  (1-e^{-s})\sum_{j\ge1}j\norm{f_j}_2^2
  =
  (1-e^{-s})\norm{\nabla f}_2^2.
\]
Finally, $1-e^{-s}\le s$, and
$\norm{\nabla h}_2\le1$ for $1$-Lipschitz $h$.
\end{proof}

\subsection{Uniform convergence for real-valued classes}
\label{sec:real-valued-uniform-convergence}

\begin{definition}[Pseudo-dimension]
    The \emph{pseudo-dimension} $\operatorname{Pdim}$ of a real-valued class $\mathcal F$
on a domain $\mathcal Z$ is the VC dimension of its subgraph class
\[
  \bigl\{
    \{(z,\tau)\in\mathcal Z\times\R:f(z)>\tau\}
    :f\in\mathcal F
  \bigr\}.
\]
\end{definition}

Equivalently, it is the largest number of inputs that can be
arbitrarily labeled by comparing a member of $\mathcal F$ with a
separate, fixed threshold at each input.
We use two standard facts: polynomial descriptions bound
pseudo-dimension, and bounded pseudo-dimension gives uniform
convergence of empirical means.

\begin{fact}[Goldberg--Jerrum, Theorem~2.2
  \cite{GoldbergJerrum1995}]
\label{fact:goldberg-jerrum}
Let
$\mathcal F=\{f_\theta:\theta\in\Theta\subseteq\R^p\}$
be a class of real-valued functions on $\mathcal Z\subseteq\R^m$.
Suppose the predicate $f_\theta(z)>\tau$ can be expressed as a
Boolean formula involving at most $S$ distinct polynomial
equalities or inequalities, each of degree at most $q$ in
$(\theta,z,\tau)$, where $q,S\ge1$. Then
\[
  \operatorname{Pdim}(\mathcal F)
  \le
  2p\log_2(8eqS).
\]
\end{fact}

The threshold $\tau$ is an input to the subgraph class, so it does
not add to the number $p$ of real parameters.

\begin{fact}[Bounded uniform convergence]
\label{fact:bounded-uniform-convergence}
Let $\mathcal F$ be a class of measurable functions taking values in
$[-M,M]$, with pseudo-dimension at most $V\ge1$. Suppose the suprema
below are measurable in the completion of the sample distribution.
For $0<\eta\le M$ and $0<\delta<1$, if
\[
 n\ge C\frac{M^2}{\eta^2}
 \left(V\log\frac{CM}{\eta}+\log\frac2\delta\right),
\]
then independent samples $\bz_1,\ldots,\bz_n$ from any distribution
satisfy, with probability at least $1-\delta$,
\[
 \sup_{f\in\mathcal F}
 \left|\frac1n\sum_{i=1}^n f(\bz_i)-\E f(\bz)\right|\le\eta.
\]
Here $C$ is a universal constant.
\end{fact}
\begin{proof}
Round each function down to a grid on $[-M,M]$ with spacing at most
$\eta/4$. Each rounded function is a constant plus a weighted sum of
$O(M/\eta)$ threshold indicators, with total weight at most $2M$.
For each fixed threshold, the indicator class has VC dimension at most
$V$. The VC uniform-convergence bound
\cite[Section~4]{BousquetBoucheronLugosi2004}, with accuracy
$\eta/(4M)$ and a union bound over the grid thresholds, gives the
stated sample bound. The rounded empirical and population means then
differ by at most $\eta/2$; the two rounding errors contribute at most
$\eta/2$ more.
\end{proof}

\section{Learning a polynomial with small squared defect}
\label{sec:polynomial-learning}

As described in \Cref{sec:proof-overview}, the first step is to learn
a low-degree polynomial with both small absolute loss and
very small squared gradient defect. We recall the definition of
squared defect for convenience:
\[
  \Gamma(p)
  :=
  \E\bigl[(\norm{\nabla p(\bx)}_2-1)_+^2\bigr].
\]
The goal of this section is to show that a polynomial satisfying
these constraints is efficiently learnable via a convex program.
We first show that a polynomial satisfying both guarantees exists,
then analyze the algorithm that learns one. The key point is that the
squared defect can be made exponentially small in the degree,
which will allow us to control the error introduced by the
subsequent Lipschitz correction.

\subsection{Existence of a low-degree witness}
\begin{lemma}[Low-degree approximation with small squared defect]
\label{lem:low-degree-witness}
Let $g$ be $1$-Lipschitz with $\norm g_2\le B$.  For a smoothing time $t>0$, let
\[
Q=(U_tg)^{\le k}.
\]
Then
\[
  \norm Q_2\le B,
  \qquad
  \norm{Q-g}_2
  \le
  \sqrt t+\frac{e^{-t(k+1)}}{\sqrt{k+1}},
  \qquad
  \Gamma(Q)\le e^{-2t(k+1)}.
\]
\end{lemma}

\begin{proof}
Write $g=\sum_{j\ge0}g_j$ and $u=U_tg$. Since $g$ is
$1$-Lipschitz,
\[
  \sum_{j\ge1}j\norm{g_j}_2^2=\norm{\nabla g}_2^2\le1.
\]
Using $(1-e^{-v})^2\le v$ gives
\[
\begin{aligned}
  \norm{g-u}_2^2
  &=\sum_{j\ge1}(1-e^{-tj})^2\norm{g_j}_2^2\le t,\\
  \norm{u-Q}_2^2
  &=\sum_{j\ge k+1}e^{-2tj}\norm{g_j}_2^2
    \le\frac{e^{-2t(k+1)}}{k+1}.
\end{aligned}
\]
Since $u$ is $1$-Lipschitz,
\[
  \Gamma(Q)
  \le\norm{\nabla(Q-u)}_2^2
  =\sum_{j\ge k+1}je^{-2tj}\norm{g_j}_2^2
  \le e^{-2t(k+1)}.
\]
The approximation bound follows by the triangle inequality,
and $\norm Q_2\le B$ since $U_t$ and truncation to 
degree $k$ do not increase $L_2$ norm.
\end{proof}

The important feature of \Cref{lem:low-degree-witness} is the
exponential decay of the squared defect with $k$.  Thus achieving
squared defect $\alpha^2$ costs only logarithmically in
$1/\alpha$ in the degree.

In the bounded realizable setting, apply the lemma to the target $g$, for which
$\by=g(\bx)$ and $\norm g_2\le B$. If $\norm{Q-g}_2\le b$, then
\begin{equation}\label{eq:witness-risk-transfer}
 \err_1(Q)=\norm{Q-g}_1\le b.
\end{equation}

\subsection{Learning the polynomial}

We now show how to learn a polynomial satisfying the guarantees
of the preceding lemma. We minimize prediction loss subject to
a constraint on squared defect and a bound on the polynomial's
$L_2$ norm. Both the objective and the constraints are
convex in the polynomial coefficients, allowing us to solve
the resulting optimization problem using samples.

The following uniform convergence result allows us to enforce
the squared-defect constraint empirically.

\begin{proposition}[Uniform convergence of squared defect]
\label{prop:uniform}
Let $\bs$ consist of $N$ independent samples from $\gauss$.
For $A\ge 1$ and $\beta,\delta\in(0,1/2]$, suppose
\[
N\ge
\left(\frac{A(d+1)^k}{\beta\delta}\right)^{10}.
\]
Then, with probability at least $1-\delta$, every polynomial
$p$ of degree at most $k$ satisfying $\norm p_2^2\le A$
simultaneously satisfies
\[
\left|
\frac1N\sum_{x\in \bs}
(\norm{\nabla p(x)}_2-1)_+^2
-\Gamma(p)
\right|
\le\beta.
\]
\end{proposition}

The proof is given in
\Cref{sec:appendix-uniform-convergence}.
In particular, taking $\beta$ sufficiently smaller than
$\alpha^2$ allows us to enforce the population squared-defect
constraint using its empirical counterpart.

\Cref{alg:learn-polynomial} states the constrained program in terms
of population statistics. Its sample implementation, analyzed in
\Cref{sec:appendix-learning-polynomial}, uses empirical penalties and
projected subgradient descent. The budget $\alpha^2/2$ reserves slack:
the implementation returns $\Gamma(p)\le\alpha^2$ with the loss
guarantee in \Cref{prop:polynomial-learning}.

\begin{algorithm}[H]
\caption{\textsc{LearnPolynomial}$(B,\eta,\alpha,\zeta)$}
\label{alg:learn-polynomial}
\begin{algorithmic}[1]
\Require Sample access to $(\bx,\by)$ with $\bx\sim\gauss$, $|\by|\le B$;
$B\ge1$; $0<\eta,\alpha,\zeta\le1$.
\Ensure A degree-$\le k$ polynomial $p$, represented by its Hermite coefficients.
\State Choose $k$ as in \eqref{eq:degree-choice} with parameters $\eta/2,\alpha/2$.
\Statex \textbf{Population program.} Minimize absolute loss over degree-$\le k$ polynomials:
\begin{equation}\label{eq:algorithm-polynomial-program}
 \begin{aligned}
 \underset{p}{\operatorname{minimize}}\quad&\err_1(p)\\
 \text{subject to}\quad&\Gamma(p)\le\alpha^2/2,\qquad \norm p_2\le B.
 \end{aligned}
\end{equation}
\State Run the sample implementation of this program from
\Cref{sec:polynomial-confidence}, with loss tolerance $\eta$
and failure probability $\zeta$.
\State \Return the Hermite coefficients of $p$.
\end{algorithmic}
\end{algorithm}

\begin{proposition}[Polynomial learning]
\label{prop:polynomial-learning}
Suppose $\bx\sim\gauss$ and $\by=g(\bx)$ for a $1$-Lipschitz function
$g:\R^d\to[-B,B]$, where $B\ge1$ is known.
For $0<\eta,\alpha,\zeta\le1$, there is a randomized algorithm that,
with probability at least $1-\zeta$, returns a degree-$k$ polynomial
$p$ satisfying
\[
  \err_1(p)\le\eta,
  \qquad
  \Gamma(p)\le\alpha^2,
  \qquad
  \norm p_2\le B,
\]
where $k$ may be chosen as in \Cref{eq:degree-choice} with parameters $\eta/2,\alpha/2$. In particular,
\[
  k
  =
  O\!\left(
    \frac{B^2}{\eta^2}
    \log\frac{B+1}{\alpha\eta}
  \right).
\]
The sample complexity and arithmetic running time are at most
\[
 P\left(B,\eta^{-1},\alpha^{-1},k,D,9^k\right)
 \left(\log\frac2\zeta\right)^2,
 \qquad D=\binom{d+k}{k},
\]
for a universal polynomial $P$.
\end{proposition}

\begin{proof}
We first analyze the base sample implementation of \Cref{alg:learn-polynomial}.
The confidence amplification in \Cref{lem:polynomial-confidence}
then gives the stated resource bound using tighter accuracy and
defect parameters.
By \Cref{lem:low-degree-witness}, the choice of $k$ gives a polynomial
with absolute loss at most $\eta/2$, squared defect at most
$\alpha^2/4$, and norm at most $B$.

The implementation minimizes empirical loss plus a penalty for exceeding
the empirical defect budget $\alpha^2/2$. Uniform convergence ensures
that the witness incurs no penalty. Comparing the empirical objective
with this witness controls the prediction loss, while comparison with
the zero polynomial bounds the defect penalty. Transferring both
bounds to the population gives $\err_1(p)\le\eta$ and
$\Gamma(p)\le\alpha^2$.

The penalized objective is convex in the Hermite coefficients.
Projected subgradient descent only requires projection onto the
coefficient ball $\norm p_2\le B$, which is explicit.
The full proof, including the sample complexity, degree, and
running-time bounds, is given in
\Cref{sec:appendix-learning-polynomial}; the general agnostic statement there
specializes to $\ell=1$ (for $L_1$ error) and $\OPT_1=0$ (the realizable setting).
\end{proof}

\section{From small squared defect to a nearby Lipschitz function}
\label{sec:hidden-witness}

As we have argued in the previous section, the polynomial learner
gives a predictor with small squared defect. We now show that
smoothing and clipping this predictor to $[-B,B]$ produces a function close in
$L_0$ to a bounded $1$-Lipschitz witness ---i.e. it agrees with a single global $B$-bounded Lipschitz function 
with high probability over a Gaussian input. 

Boundedness will make local Lipschitz bounds sufficient. If a
function takes values in $[-B,B]$, then points at distance at least
$2B$ automatically satisfy the Lipschitz inequality. We therefore
only need to control nearby points. Harnack's inequality gives this
control around most Gaussian points, and a McShane envelope then
produces a globally $1$-Lipschitz function that agrees with the
predictor at those points.

Importantly, the algorithm does not compute the McShane envelope.
It is just an analytical tool to show that the (smoothed, clipped) output of the 
polynomial learner is $L_0$-close to ${\cal{L}}_{B}$.
Fix $s>0$ and write
\[
  P=U_sp,\qquad q=\cB(P),\qquad
  r=(\norm{\nabla p}-1)_+,
\]
where $\cB$ clips to $[-B,B]$. By definition,
$\E r(\bx)^2=\Gamma(p)$. Since $\norm{\nabla p}\le1+r$,
the OU derivative identity gives
\begin{equation}\label{eq:smoothed-gradient}
  \norm{\nabla P(z)}
  \le \rho_s U_s\norm{\nabla p}(z)
  \le \rho_s (1+U_sr(z)).
\end{equation}
In particular, wherever $U_sr(z)\le e^s-1$, we have
$\norm{\nabla P(z)}\le1$.

\begin{lemma}[Agreement with a bounded Lipschitz witness]
\label{lem:smoothed-bridge}
The McShane envelope
\[
  h(x)=\inf_{y\in\R^d}\{q(y)+\norm{x-y}\}
\]
belongs to $\calL_{B}$ and satisfies
\begin{equation}\label{eq:mcshane-agreement}
  \Pr[q(\bx)\ne h(\bx)]
  \le \Gamma(p) \cdot C_{s,B},\qquad 
  C_{s,B}=
  \frac{1}{(e^s-1)^2}
  \exp\!\left(\frac{4B^2}{e^{2s}-1}\right).
\end{equation}
Additionally,
\[\norm{q-h}_1 \le 
  \norm{q-h}_2
  \le 2B\sqrt{\Gamma(p) \cdot C_{s,B}},
\]
\end{lemma}

\begin{proof}
By Wang's dimension-free Harnack inequality \cite{Wang1997},
for every nonnegative function $u$ and every $x,z\in\R^d$,
\begin{equation}\label{eq:ou-likelihood}
  U_su(z)
  \le\sqrt{U_s(u^2)(x)}
  \exp\!\left(\frac{\norm{z-x}^2}{2(e^{2s}-1)}\right).
\end{equation}
Set
\[
  a=(e^s-1)
  \exp\!\left(-\frac{2B^2}{e^{2s}-1}\right),
  \qquad
  S=\{x:U_s(r^2)(x)\le a^2\}.
\]
For $x\in S$ and $\norm{z-x}\le2B$,
\Cref{eq:ou-likelihood} gives
\[
  U_sr(z)
  \le\sqrt{U_s(r^2)(x)}
  \exp\!\left(\frac{2B^2}{e^{2s}-1}\right)
  \le e^s-1.
\]
By \Cref{eq:smoothed-gradient}, $\norm{\nabla P}\le1$
throughout the radius-$2B$ ball around $x$. Integrating along
the segment from $x$ to $y$, and then clipping, gives
\[
  |q(x)-q(y)|\le\norm{x-y}
  \qquad\text{if }\norm{x-y}\le2B.
\]
For points outside this ball, boundedness gives the same inequality:
$|q(x)-q(y)|\le2B<\norm{x-y}$.
Thus every $x\in S$ satisfies
\begin{equation}\label{eq:q-good-center}
  q(x)\le q(y)+\norm{x-y}
  \qquad\text{for every }y\in\R^d.
\end{equation}

The McShane envelope $h$
is $1$-Lipschitz, and additionally $-B\le h\le q\le B$.
For $x\in S$, \Cref{eq:q-good-center} gives $h(x)\ge q(x)$;
the candidate $y=x$ gives the reverse inequality. Hence $h=q$ on $S$.

Gaussian stationarity and Markov's inequality now give
\[
  \Pr[q(\bx)\ne h(\bx)]
  \le\gauss(\R^d \setminus S)
  \le\frac{\E U_s(r^2)(\bx)}{a^2}
  =\frac{\Gamma(p)}{a^2}
  =\Gamma(p)C_{s,B},
\]
which is \Cref{eq:mcshane-agreement}. Since Gaussian measure is a
probability measure, Cauchy--Schwarz gives $\norm{q-h}_1\le\norm{q-h}_2$.
Combining this with $|q-h|\le2B$ yields
\[
  \norm{q-h}_1\le\norm{q-h}_2
  \le2B\sqrt{\Pr[q(\bx)\ne h(\bx)]}
  \le2B\sqrt{\Gamma(p)C_{s,B}}.
\]
\end{proof}

\subsection{Consequences for the learned predictor}

By nonexpansiveness of clipping and \Cref{eq:ou-smoothing-cost},
\begin{equation}\label{eq:smoothing-slack}
  \norm{q-\cB(p)}_2
  \le\norm{U_sp-p}_2
  \le\sqrt{s}(1+\sqrt{\Gamma(p)}).
\end{equation}
Thus smoothing changes the predictor only slightly, while
\Cref{lem:smoothed-bridge} gives a bounded $1$-Lipschitz witness
that agrees with $q$ outside a small set. In particular, choosing
\[
  \Gamma(p)\le\frac{\omega^2}{4B^2C_{s,B}}
\]
ensures that $q$ is within $\omega$ in $L_2(\gauss)$ of
$\calL_B$.

The witness is not computed. In the next section, we use its
agreement with $q$ to analyze a local Gaussian correction that
can be approximated using finitely many evaluations of the polynomial.

\section{From small squared defect to pointwise closeness}
\label{sec:ideal-repair}

In the previous section, we showed that smoothing and clipping a
polynomial $p$ with small squared defect gives a function $q$
that is close to a McShane witness $h$ in $L_2$ norm and
agrees with it outside a small set. We now use Gaussian probes
to turn this agreement into an efficient local correction
algorithm.
The algorithm's output $G$ is close to $p$ in $L_1$ norm
and uniformly close to a single $1$-Lipschitz function $H$,
simultaneously at every query. All norms and probabilities below
are with respect to $\gauss$, unless otherwise specified.

\begin{proposition}[Local correction with a pointwise guarantee]
\label{prop:pointwise-correction}
Let $p:\R^d\to\R$ be a polynomial of degree at most $k$, and let
$0<\beta\le\varepsilon\le1/4$ and $0<\delta\le1/2$. Suppose
$\Gamma(p)\le\exp(-4/\varepsilon^4)$.
Given the center $c=\E p$, there is a randomized local computation
algorithm with oracle access to $p$ that answers queries to
$G:\R^d\to[c-\varepsilon^{-1},c+\varepsilon^{-1}]$.
With probability at least $1-\delta$,
\[
  \norm{G-p}_1\le31\varepsilon,\qquad
  \norm{G-c}_1\le20,
\]
and there is a single $1$-Lipschitz function $H$ such that
\[
  |G(x)-H(x)|\le\beta\qquad\text{for every }x\in\R^d.
\]
All randomness is fixed during preprocessing, so the guarantees
hold simultaneously for every query. The algorithm uses at most
\[
  (d+1)^{C(k+\varepsilon^{-4})}\beta^{-C}\log(2/\delta)
\]
evaluations per query, for a universal constant $C$.
\end{proposition}

\subsection{Centering at the mean}

We will later clip the smoothed polynomial at height
$L=\varepsilon^{-1}$. To bound the error from clipping, we need
a bound on the second moment. Small squared defect controls
the variance of $p$, but not its mean, so we subtract $c=\E p$.
The Gaussian Poincar\'e inequality gives
\[
  \norm{p-c}_2
  =\sqrt{\operatorname{Var}(p)}
  \le\norm{\nabla p}_2
  \le1+\sqrt{\Gamma(p)}\le2.
\]
The corrector takes $c$ as an input. In the learning algorithm,
$p$ is represented in the Hermite basis, so $c$ is its constant
Hermite coefficient and requires no estimation.

\subsection{The ideal Gaussian corrector}

We would like to use the McShane witness $h$ as our corrected
function, but we do not have an efficient way to evaluate it,
as we cannot exactly compute distance to sublevel sets. We can,
however, estimate Gaussian measures of the sublevel sets by sampling.
We convert these Gaussian measures into a penalty term that can be 
used as a proxy for distance: a rare sublevel set is treated as if 
it is far away. 

We show three properties of this corrector: 
\begin{itemize}
\item[(a)] It always returns a Lipschitz function.
\item[(b)] When run on a function in ${\cal{L}}_L$, it is equivalent to \emph{median smoothing}, which doesn't introduce much error.
\item[(c)] It is stable under changes on a small set; if $\Pr[f(\bx) \ne g(\bx)]$ is small, then $\norm{\repair[f] - \repair[g]}_1$ is small.
\end{itemize}

Fix $L>0$ and a smoothing time $s>0$, and let
\[
  c_s=\frac{\sigma_s}{\rho_s}=\sqrt{e^{2s}-1}.
\]
At each $x$, write $\bw_x=\rho_s x+\sigma_s \bz$, where $\bz\sim\gauss$.
For measurable $v:\R^d\to[-L,L]$, define
\[
  q_a[v](x)=\Pr_{\bz} [v(\bw_x)\le a].
\]
Thus $q_a[v](x)$ is the probability that a Gaussian probe
at $x$ returns a value at most $a$. The lower median
is the smallest threshold for which this probability reaches
$1/2$.

Let $\Phi$ and $\varphi$ be the standard Gaussian CDF
and density. Define the penalty
\[
  \psi(r)=\min\{2L,c_s[-\Phi^{-1}(r)]_+\}
  \quad(0<r<1),\qquad
  \psi(0)=2L,\quad\psi(1)=0.
\]
The inverse Gaussian CDF expresses the vote probability on
a Gaussian quantile scale. The penalty is zero for $r\ge1/2$
and increases as $r$ decreases, up to the cap $2L$;
see \Cref{fig:gaussian-corrector}(a).

The corrected value is the smallest penalized threshold:
\begin{definition}[Ideal corrector]\label{def:ideal-corrector}
For a function $v$, define the ideal corrector as follows:
\begin{equation}\label{eq:ideal-repair}
  \repair[v](x)=\inf_{a\in[-L,L]}
  \{a+\psi(q_a[v](x))\}.
\end{equation}
\end{definition}

The output lies in $[-L,L]$: every score is at least $-L$,
and the threshold $L$ gives score $L$. Capping the penalty
does not change the infimum, since any capped score is
already at least $L$.

By the Gaussian smoothing inequality, the choice of
$\Phi^{-1}$ and the factor $c_s$ makes each threshold score
$1$-Lipschitz in $x$. We prove this first.

For a $1$-Lipschitz input, the infimum equals the Gaussian
median. The median threshold has zero penalty. Raising the
threshold increases its value without reducing the penalty;
lowering it costs at least as much in penalty as it saves
in threshold value. This is illustrated in
\Cref{fig:gaussian-corrector}(b) and proved below.

\begin{proposition}[Ideal properness]
\label{prop:ideal-properness}
For every measurable $v:\R^d\to[-L,L]$, the function
$\repair[v]$ is $1$-Lipschitz.
\end{proposition}

\begin{proof}
We show that each threshold score
\[
  x\mapsto a+\psi(q_a[v](x))
\]
is $1$-Lipschitz.

Fix $a$. If $\{v\le a\}$ has Gaussian measure zero or one,
then $q_a[v]$ is constant, so the claim holds. Otherwise,
the Gaussian smoothing inequality \cite{BakryGentilLedoux2015} gives
\[
  \left|\Phi^{-1}(q_a[v](x))-\Phi^{-1}(q_a[v](y))\right|
  \le \frac{\rho_s}{\sigma_s}\norm{x-y}.
\]
Multiplying by $c_s=\sigma_s/\rho_s$ gives Lipschitz constant
one. Taking the negative part and capping at $2L$ cannot
increase it. Thus
\[
  |\psi(q_a[v](x))-\psi(q_a[v](y))|
  \le \norm{x-y}.
\]
Adding the constant $a$ preserves this bound. Finally,
$\repair[v]$ is the infimum of these scores, so it satisfies
the same Lipschitz bound.
\end{proof}

We now prove that, for a $1$-Lipschitz input $v$, the corrected 
function equals the median of $v(\rho_s x+\sigma_s \bz)$ and is close to $v$
in $L_2$ norm.

\begin{lemma}[Corrector computes the Gaussian median]
\label{lem:repair-median}
Let $v:\R^d\to[-L,L]$ be $1$-Lipschitz, and let $m_v(x)$
be the lower median of $v(\rho_s x+\sigma_s \bz)$. Then
\[
  \repair[v](x)=m_v(x)
  \qquad\text{for every }x\in\R^d.
\]
\end{lemma}

\begin{proof}
Fix $x$ and write $m=m_v(x)$. The threshold $m$ has at least
half the votes, so its penalty is zero and its score is $m$.
We show that no other threshold has a smaller score.

For $a>m$, the score is at least $a>m$, since the penalty
is nonnegative. Now fix $a<m$. We first show that
\[
  q_a[v](x)\le\Phi\left(\frac{a-m}{\sigma_s}\right).
\]
Set $f(z)=v(\rho_s x+\sigma_s z)$, so $f$ is $\sigma_s$-Lipschitz,
and let $A=\{z:f(z)\ge m\}$. Since $m$ is a median,
$\Pr[\bz\in A]\ge1/2$. If $f(z)\le a$ and $w\in A$, then
\[
  m-a\le f(w)-f(z)\le\sigma_s\norm{w-z}.
\]
Thus every point of $\{f\le a\}$ is at distance at least
$(m-a)/\sigma_s$ from $A$. For $0<r<(m-a)/\sigma_s$, let
\[
  A_r=\{z:\operatorname{dist}(z,A)<r\}.
\]
Gaussian isoperimetry gives $\Pr[\bz\in A_r]\ge\Phi(r)$.
Since $\{f\le a\}$ is disjoint from $A_r$,
\[
  q_a[v](x)\le1-\Phi(r).
\]
Letting $r\uparrow(m-a)/\sigma_s$ proves the claimed bound.

If $q_a[v](x)=0$, its penalty is $2L\ge m-a$.
Otherwise, the bound above gives
\[
  c_s[-\Phi^{-1}(q_a[v](x))]_+
  \ge \frac{m-a}{\rho_s}
  \ge m-a.
\]
Since $m-a\le2L$, capping the penalty at $2L$ preserves
this lower bound. Therefore, the score at $a$ is at least
$a+(m-a)=m$.
\end{proof}

\begin{lemma}[Corrector stays close to a Lipschitz input]
\label{lem:repair-lipschitz-accuracy}
If $v:\R^d\to[-L,L]$ is $1$-Lipschitz, then
\[
  \norm{\repair[v]-v}_2\le2\sqrt{1-\rho_s}.
\]
\end{lemma}

\begin{proof}
By \Cref{lem:repair-median}, it suffices to bound
$\norm{m_v-v}_2$.

For any real random variable $\bv$, any median $m$, and any
$t\in\R$,
\[
  |m-t|^2\le2\E|\bv-t|^2.
\]
Indeed, with probability at least $1/2$, $\bv$ lies on the side
of $m$ away from $t$, where $|\bv-t|\ge|m-t|$.
Applying this with $\bv=v(\rho_s x+\sigma_s \bz)$ and $t=v(x)$ gives
\[
  |m_v(x)-v(x)|^2
  \le2\E_{\bz}|v(\rho_s x+\sigma_s \bz)-v(x)|^2.
\]
Let $\bx,\bz$ be independent standard Gaussian vectors and set
$\by=\rho_s \bx+\sigma_s \bz$. Integrating over $x=\bx$, we obtain
\[
  \norm{m_v-v}_2^2
  \le2\E|v(\by)-v(\bx)|^2.
\]

To bound the right-hand side, write $v=\sum_{j\ge0}v_j$,
where $v_j$ is its degree-$j$ Hermite component.
The Gaussian noise and gradient identities give
\[
\begin{aligned}
  \E|v(\by)-v(\bx)|^2
  &=2\sum_{j\ge1}(1-\rho_s^j)\norm{v_j}_2^2,\\
  \norm{\nabla v}_2^2
  &=\sum_{j\ge1}j\norm{v_j}_2^2.
\end{aligned}
\]
Since
\[
  1-\rho_s^j
  =(1-\rho_s)(1+\rho_s+\cdots+\rho_s^{j-1})
  \le j(1-\rho_s),
\]
it follows that
\[
  \E|v(\by)-v(\bx)|^2
  \le2(1-\rho_s)\norm{\nabla v}_2^2.
\]
Combining the bounds and using $\norm{\nabla v}_2\le1$,
we conclude that
\[
  \norm{m_v-v}_2^2
  \le4(1-\rho_s)\norm{\nabla v}_2^2
  \le4(1-\rho_s).
\]
Taking square roots proves the claim.
\end{proof}

\subsection{Only bad probes can change the answer}

We want to show that $\repair[q]$ is close to the McShane
witness $h$. We do this in two steps: compare $\repair[q]$
to $\repair[h]$, then compare $\repair[h]$ to $h$.
The previous lemma gives the second comparison.
For the first, we use the fact that $q$ and $h$ agree
outside a small set. A Gaussian probe outside this set
gives the same vote for both functions, so only probes
that hit the disagreement set can affect the comparison.

To turn a bound on vote probabilities into a bound on the
corrected values, we need to control the sensitivity of
$\psi$.
Without the cap, the penalty becomes arbitrarily steep as
$r\to0$. The cap replaces this part by a constant.
To see the resulting Lipschitz bound, put
\[
  r_0=\Phi(-2L/c_s).
\]
Then $\psi(r)=2L$ for $r\le r_0$ and $\psi(r)=0$ for
$r\ge1/2$. On the remaining interval,
\[
  |\psi'(r)|
  =\frac{c_s}{\varphi(\Phi^{-1}(r))}
  \le\frac{c_s}{\varphi(2L/c_s)}
  =:K,
\]
since $\Phi^{-1}(r)\in[-2L/c_s,0]$.
Thus $\psi$ is continuous, has slope at most $K$ in
magnitude on its decreasing part, and is constant elsewhere.
It is therefore $K$-Lipschitz, with
\[
  K=c_s\sqrt{2\pi}\,e^{2L^2/c_s^2}.
\]

The resulting stability bound loses a factor $K$, which is
exponential in $L^2/c_s^2$. For our later choices
$L=\varepsilon^{-1}$ and $s=2\varepsilon^2$, this factor
is $\exp(O(\varepsilon^{-4}))$. We can afford this loss: the small squared defect assumption,
together with the preceding smoothing argument, makes the
disagreement probability small enough to compensate for $K$. We check this balance in the next
subsection.

\begin{lemma}[Stability under changes on a small set]
\label{lem:hard-repair-stability}
For measurable $v_1,v_2:\R^d\to[-L,L]$,
\[
  \norm{\repair[v_1]-\repair[v_2]}_1
  \le K\Pr[v_1(\bx)\ne v_2(\bx)].
\]
\end{lemma}

\begin{proof}
Let $E=\{y:v_1(y)\ne v_2(y)\}$. Fix a query $x$ and
use the same probe $\bw_x=\rho_s x+\sigma_s \bz$ for both functions.
Recall that $q_a[v_i](x)=\Pr[v_i(\bw_x)\le a]$.
If $\bw_x\notin E$, then $v_1(\bw_x)=v_2(\bw_x)$. Hence, for
any threshold $a$, either both values are at most $a$ or
neither is. The two indicators below can therefore differ
only when $\bw_x\in E$.

Split each probability according to whether the probe lands
inside or outside $E$:
\[
  q_a[v_i](x)
  =\Pr[\bw_x\notin E,\ v_i(\bw_x)\le a]
   +\Pr[\bw_x\in E,\ v_i(\bw_x)\le a],
  \qquad i=1,2.
\]
Since $v_1=v_2$ outside $E$, the first terms are equal
and cancel when we subtract. Hence
\[
\begin{aligned}
  |q_a[v_1](x)-q_a[v_2](x)|
  &=\bigl|
    \Pr[\bw_x\in E,\ v_1(\bw_x)\le a]\\
  &\hspace{2.5em}
    -\Pr[\bw_x\in E,\ v_2(\bw_x)\le a]
    \bigr|\\
  &\le\Pr[\bw_x\in E].
\end{aligned}
\]
The last inequality holds because both probabilities lie
between $0$ and $\Pr[\bw_x\in E]$, so their difference in
absolute value is at most the length of this interval.
This bound holds for every threshold $a$.
Since $\psi$ is $K$-Lipschitz, the corresponding threshold
scores differ by at most $K\Pr[\bw_x\in E]$.

Taking infima over $a$ preserves this bound, giving
\[
  |\repair[v_1](x)-\repair[v_2](x)|
  \le K\Pr[\bw_x\in E].
\]
Finally, integrate over an independent $\bx\sim\gauss$.
Since $\rho_s \bx+\sigma_s \bz$ is standard Gaussian,
\[
  \norm{\repair[v_1]-\repair[v_2]}_1
  \le K\Pr[\rho_s \bx+\sigma_s \bz\in E]
  =K\Pr[\bx\in E].
\]
\end{proof}

\subsection{Accuracy after smoothing}

Fix $0<\varepsilon\le1/4$ and choose
\[
  L=\varepsilon^{-1},\qquad s=2\varepsilon^2,
\]
and set $Q=U_s(p-c)$ and $q=C_L(Q)$. We use this same time $s$
for the Gaussian probes in $\repair$. Since $0<\varepsilon\le1/4$,
\[
  \varepsilon\le\sigma_s\le2\varepsilon\le\tfrac12,
  \qquad 1-\rho_s\le s=2\varepsilon^2.
\]
We now combine the preceding bounds. The McShane witness $h$
agrees with $q$ outside a small set, so $\repair[q]$ is close
to $\repair[h]$ by stability. Since $\repair[h]$ is close to
$h$, and $h$ agrees with $q$ outside that small set, we obtain
a bound on the error introduced by repairing $q$.

\begin{proposition}[Accuracy after smoothing]
\label{prop:ideal-accuracy}
If $\norm{p-c}_2<10$ and $\Gamma(p)\le e^{-4/\varepsilon^4}$,
then
\[
  \norm{\repair[q]-(p-c)}_1\le30\varepsilon.
\]
\end{proposition}

\begin{proof}
Apply \Cref{lem:smoothed-bridge} to $p-c$, whose squared
defect is $\Gamma(p)$. It gives $h\in\calL_{L}$ with
\[
  \mu:=\Pr[q(\bx)\ne h(\bx)]
  \le\frac{\Gamma(p)}{(e^s-1)^2}
    \exp\!\left(\frac{4L^2}{e^{2s}-1}\right)
  \le\frac1{4\varepsilon^4}e^{-3/\varepsilon^4}.
\]
The last inequality uses $e^u-1\ge u$ and the stated parameters.

Compare $q$ to its repair through $h$ and $\repair[h]$.
By \Cref{lem:hard-repair-stability,lem:repair-lipschitz-accuracy}
and $\norm{q-h}_1\le2L\mu$,
\[
\begin{aligned}
  \norm{\repair[q]-q}_1
  &\le\norm{\repair[q]-\repair[h]}_1
      +\norm{\repair[h]-h}_1+\norm{h-q}_1\\
  &\le K\mu+2\sqrt{1-\rho_s}+2L\mu.
\end{aligned}
\]
Since $\sigma_s\le1/2$, we have $c_s\le2\sigma_s/\sqrt3$ and hence
\[
  K\le3\sigma_s e^{2L^2/\sigma_s^2}
  \le6\varepsilon e^{2/\varepsilon^4}.
\]
Together with the bound on $\mu$, this gives
$(K+2L)\mu\le\varepsilon/2$ for $0<\varepsilon\le1/4$.
Also, $2\sqrt{1-\rho_s}\le2\sqrt2\,\varepsilon\le3\varepsilon$.
Hence $\norm{\repair[q]-q}_1\le\tfrac72\varepsilon$.

It remains to bound the error from smoothing and clipping.
Recall that $Q=U_s(p-c)$ and $q=C_L(Q)$. We compare
$p-c$ to $Q$ first, then $Q$ to $q$:
\[
  \norm{q-(p-c)}_1
  \le\norm{Q-(p-c)}_1+\norm{q-Q}_1.
\]

For the smoothing error, we use the bound
\[
  \norm{U_s(p-c)-(p-c)}_2
  \le\sqrt{s}\,\norm{\nabla p}_2,
\]
Since the Gaussian measure is a probability measure, this gives
\[
  \norm{Q-(p-c)}_1
  \le\norm{Q-(p-c)}_2
  \le\sqrt{s}\,\norm{\nabla p}_2.
\]
Since $s=2\varepsilon^2$ and
\[
  \norm{\nabla p}_2
  \le1+\sqrt{\Gamma(p)}
  \le1+e^{-2/\varepsilon^4},
\]
we obtain, for $0<\varepsilon\le1/4$,
\[
  \norm{Q-(p-c)}_1
  \le\sqrt{2}\,\varepsilon
       \bigl(1+e^{-2/\varepsilon^4}\bigr)
  \le\frac32\varepsilon.
\]

For the clipping error, clipping changes $Q(x)$ only when
$|Q(x)|>L$, and the amount removed is exactly $|Q(x)|-L$.
Thus
\[
  \norm{q-Q}_1=\E\bigl[(|Q(\bx)|-L)_+\bigr].
\]
The pointwise inequality
\[
  (|t|-L)_+\le\frac{t^2}{4L}
\]
bounds this error by the second moment. For $|t|>L$, it follows
from $t^2-4L(|t|-L)=(|t|-2L)^2\ge0$; for $|t|\le L$,
the left-hand side is zero. Consequently,
\[
  \norm{q-Q}_1\le\frac{\norm Q_2^2}{4L}.
\]
Smoothing cannot increase the $L_2$ norm, so
$\norm Q_2\le\norm{p-c}_2<10$. With $L=\varepsilon^{-1}$,
this gives
\[
  \norm{q-Q}_1\le\frac{100}{4L}=25\varepsilon.
\]

Adding the smoothing and clipping errors, we conclude that
\[
  \norm{q-(p-c)}_1
  \le\frac32\varepsilon+25\varepsilon.
\]
Together with the preceding bound
$\norm{\repair[q]-q}_1\le\frac72\varepsilon$, this yields
\[
  \norm{\repair[q]-(p-c)}_1
  \le\frac72\varepsilon+\frac32\varepsilon+25\varepsilon
  \le30\varepsilon.
\]
\end{proof}

\subsection{Local implementation of the corrector}

\Cref{alg:phase-one} separates preprocessing from query evaluation.
Returning a function means returning its evaluation procedure together
with the stored probes, as in the succinct-representation convention
of~\cite[Section~2.4]{LV2023}. The center $c=\E p$ is supplied,
and all randomness comes from the Gaussian probes. The final learner
reads $c$ from the constant Hermite coefficient of $p$; see
\Cref{prop:finite-repair-interface}.

\begin{algorithm}[H]
\caption{\textsc{QuantileCorrector}$(p,c,\varepsilon,\beta,\delta)$}
\label{alg:phase-one}
\begin{algorithmic}[1]
\Require Oracle access to a degree-$k$ polynomial $p$; its mean $c=\E p$;
$0<\beta\le\varepsilon\le1/4$; $0<\delta\le1/2$.
\Ensure A representation of $G$ consisting of $c$, stored probes and penalties, and the evaluation rule below.
\Statex \textbf{Preprocessing (once).}

\State Set
\[
 L=\varepsilon^{-1},\qquad s=2\varepsilon^2.
\]
Define $Q=U_s(p-c)$ and $q=C_L(Q)$.
\State Choose $N$ using the VC bound in the proof of \Cref{prop:uniform-corrector-approximation}
with failure probability $\delta$.
\State Draw and store independent $\bz_1,\ldots,\bz_N\sim\gauss$.
Compute and store $\psi(j/N)$ for $j=1,\ldots,N$.

\Statex \textbf{Evaluation at a query $x$.}

\State Compute $u_i=q(\rho_s x+\sigma_s \bz_i)$ for $i=1,\ldots,N$.
\State Sort the values as $u_{(1)}\le\cdots\le u_{(N)}$.
\State \Return
\[
 G(x)=c+\min_{1\le j\le N}\{u_{(j)}+\psi(j/N)\}.
\]

\Statex \textbf{Evaluating the smoothed polynomial.} To compute $Q(y)$, interpolate
$P_y(z)=p(\rho_s y+\sigma_s z)-c$
from $\binom{d+k}{k}$ evaluations of $p$, then compute
$Q(y)=\E[P_y(\bz)]$ using the Gaussian moments of its monomials.
\Statex Every query reuses the stored probes and penalties; evaluation uses no fresh randomness.
\end{algorithmic}
\end{algorithm}

The evaluation rule in \Cref{alg:phase-one} uses
\begin{equation}\label{eq:sampled-repair}
 \widehat{\repair}[q](x)
 =\min_{1\le j\le N}\{u_{(j)}+\psi(j/N)\}.
\end{equation}
It replaces ideal vote probabilities by empirical frequencies.
The next proposition bounds the approximation error
simultaneously for all queries.

\begin{proposition}[Uniform convergence]
\label{prop:uniform-corrector-approximation}
Let $q=C_L(Q)$ for a degree-$k$ polynomial $Q$. Suppose
$L\ge1$, $s>0$, $0<\beta\le\sigma_s\le1/2$, and $0<\delta\le1/2$.
For a universal $C$, if
\[
  N\ge(d+1)^{C(k+L^2/\sigma_s^2)}\beta^{-C}\log(2/\delta),
\]
then, with probability at least $1-\delta$,
\[
  \sup_{x\in\R^d}
  |\widehat{\repair}[q](x)-\repair[q](x)|\le\beta.
\]
\end{proposition}

\begin{proof}
Define the empirical votes
\[
  \widehat q_a(x)
  =\frac1N\sum_{i=1}^N
    \mathbf1_{\{q(\rho_s x+\sigma_s \bz_i)\le a\}}.
\]
Let $\Psi(z)$ list the $D=\binom{d+k}{k}$ monomials of
degree at most $k$, including $1$. For $a\in[-L,L)$,
\[
  Q(\rho_s x+\sigma_s z)-a=b_{x,a}\cdot\Psi(z)
\]
for some coefficient vector $b_{x,a}$. Thus the vote event is
\[
  \{z:q(\rho_s x+\sigma_s z)\le a\}
  =\{z:b_{x,a}\cdot\Psi(z)\le0\}.
\]
This is one linear inequality in the fixed feature vector
$\Psi(z)$. The vote events are therefore inverse images
of halfspaces in $\R^D$, so their VC dimension is at most
$D+1$. At $a=L$, the event is the whole space, also
represented by the constant polynomial $-1$.

Standard VC uniform convergence gives, with probability
at least $1-\delta$,
\[
  \sup_{x\in\R^d,\ a\in[-L,L]}
  |\widehat q_a(x)-q_a[q](x)|\le\beta/K,
\]
provided
\[
  N\ge C_0\frac{K^2}{\beta^2}
  \left((D+1)\log\frac{2K}{\beta}+\log\frac2\delta\right).
\]
The assumed sample bound suffices after increasing $C$, since
\[
  D\le(d+1)^k,
  \qquad K\le3\sigma_s e^{2L^2/\sigma_s^2}.
\]

On this event, compare the ideal and empirical scores
at the same threshold:
\[
  S_x(a)=a+\psi(q_a[q](x)),
  \qquad
  \widehat S_x(a)=a+\psi(\widehat q_a(x)).
\]
Since $\psi$ is $K$-Lipschitz,
\[
\begin{aligned}
  |\widehat S_x(a)-S_x(a)|
  &=|\psi(\widehat q_a(x))-\psi(q_a[q](x))|\\
  &\le K|\widehat q_a(x)-q_a[q](x)|
  \le\beta.
\end{aligned}
\]
This holds for every $x,a$. Taking infima over $a$ gives
\[
  \repair[q](x)-\beta
  \le\inf_{a\in[-L,L]}\widehat S_x(a)
  \le\repair[q](x)+\beta.
\]

We finally check that \Cref{alg:phase-one} computes this empirical
infimum using actual vote counts. If some sample value is
at most $a$, let $t$ be the largest such value. Exactly the
same samples are at most $t$ as are at most $a$, so
\[
  \widehat q_t(x)=\widehat q_a(x),
  \qquad
  \widehat S_x(t)\le\widehat S_x(a).
\]
If no sample value is at most $a$, its score is
$a+\psi(0)=a+2L\ge L$. The largest sample value has score
at most $L$, so such a threshold cannot improve the minimum.
Therefore,
\[
\begin{aligned}
  \inf_{a\in[-L,L]}\widehat S_x(a)
  &=\min_{t\in\{u_1,\ldots,u_N\}}
      \{t+\psi(\widehat q_t(x))\}\\
  &=\widehat{\repair}[q](x).
\end{aligned}
\]
Here $\widehat q_t(x)$ counts all samples at most $t$,
including repeated values. Combining this identity with
the preceding comparison proves
\[
  \sup_{x\in\R^d}
  |\widehat{\repair}[q](x)-\repair[q](x)|\le\beta.
\]
\end{proof}

We are now ready to prove \Cref{prop:pointwise-correction}.

\begin{proof}[Proof of \Cref{prop:pointwise-correction}]
Run \Cref{alg:phase-one} with the supplied center $c=\E p$ and
confidence parameter $\delta$. The centering bound above gives
$\norm{p-c}_2\le2<10$. Set
\[
  G=c+\widehat{\repair}[q],\qquad H=c+\repair[q].
\]
Here $H$ is the shifted ideal corrected output, distinct from the McShane
witness $h$. By \Cref{prop:ideal-properness}, $H$ is
$1$-Lipschitz.
By \Cref{prop:uniform-corrector-approximation}, with probability
at least $1-\delta$, $|G(x)-H(x)|\le\beta$ for every $x$.
On this event,
\Cref{prop:ideal-accuracy} and the uniform approximation give
\[
  \norm{G-p}_1\le30\varepsilon+\beta\le31\varepsilon,
  \qquad
  \norm{G-c}_1\le2+31\varepsilon<20.
\]
The range of $G$ follows from \Cref{eq:sampled-repair}.

Each query uses $N$ evaluations of $q$. To compute $q(y)$,
we interpolate the degree-$k$ polynomial
\[
  z\mapsto p(\rho_s y+\sigma_s z)-c
\]
using $D=\binom{d+k}{k}\le(d+1)^k$ evaluations of $p$.
Integrating its monomials using the Gaussian moments gives
$Q(y)$ exactly. Clipping then gives $q(y)$.
Therefore, each query uses at most $ND$ evaluations of $p$.
With $L^2/\sigma_s^2\le\varepsilon^{-4}$, the VC bound in the proof of
\Cref{prop:uniform-corrector-approximation} allows us to choose $N$ so that
\[
  ND\le(d+1)^{C(k+\varepsilon^{-4})+k}\beta^{-C}\log(2/\delta).
\]
The additional $k$ in the exponent is absorbed by increasing
the universal constant $C$, giving the stated query bound.
\end{proof}

\section{From pointwise closeness to a proper learner}
\label{sec:finite-proper}

The previous section supplied a function $G$ that is pointwise close
to a specified $1$-Lipschitz witness $H$. In this section, for simplicity
of analysis, we subtract the center from the previous section's
construction. Write
\[
 v=G-c=\widehat{\repair}[q],\qquad
 h_\star=H-c=\repair[q],\qquad r=v-h_\star,
\]
where $h_\star$ is the ideal Gaussian corrector
(\Cref{def:ideal-corrector}). The ideal corrector is $1$-Lipschitz
and takes values in $[-L,L]$. The empirical corrector $v$ also takes
values in $[-L,L]$, and, on the uniform approximation event,
$\norm r_\infty\le\beta$.

While the values of $v$ are controlled and close to those of
$h_\star$, the guarantees so far still permit steep local oscillations,
leading to violations of the Lipschitz condition. Our objective in
this section is to smooth these oscillations. To make this work
efficiently, we approximate the smoothed function and its gradient
by a finite average, and rescale to obtain a $1$-Lipschitz predictor.

Throughout this section, we work with the centered functions and
restore $c$ at the end. We prove the following.

\begin{proposition}[Proper correction]\label{prop:proper}
Let $L\ge1$, $T_0\ge1$, $0<\beta\le1/16$, and $0<\delta<1/2$.
Let $Q:\R^d\to\R$ be a polynomial of degree at most $k$ with
$\norm Q_2<10$, and let $q=C_L(Q)$ be its clipping to $[-L,L]$.
Write $h_\star=\repair[q]$ for the ideal $1$-Lipschitz corrector and
$v=\widehat{\repair}[q]$ for the empirical corrector constructed
from $N$ stored probes $\bz_1,\ldots,\bz_N$ satisfying
\[
 \norm{\bz_i}\le T_0\qquad\text{for every }i\le N.
\]

There is a randomized algorithm that, given query access to $v$
and the bound $T_0$, answers queries to a single function
$v_{\mathrm{prop}}:\R^d\to[-L,L]$.
During preprocessing, it draws and stores $M$ new independent
standard Gaussian vectors, independently of the stored probes
defining $v$. Every subsequent query uses these same new vectors and
at most $M$ evaluations of $v$, where
\begin{equation}\label{eq:complexity}
 M\le
 \left(
   \frac{(d+1)(k+1)(L+1)(1+T_0)}{\beta}
 \right)^{C(k+1)}
 \left[
   \log(N+1)+\log\frac2\delta
 \right]
\end{equation}
for a universal constant $C$.

For every fixed choice of the stored probes satisfying the norm bound,
with probability at least $1-\delta$ over the $M$ new Gaussian vectors,
the following implication holds: if $\norm{v-h_\star}_\infty\le\beta$, then
\[
 \Lip(v_{\mathrm{prop}})\le1,
 \qquad
 \norm{v_{\mathrm{prop}}-v}_2\le8(L+1)\sqrt\beta.
\]
Because all randomness is fixed during preprocessing, the guarantees
concern the entire output function and hold simultaneously for every query.
\end{proposition}

The stored probes need not be Gaussian for this proposition.
In the modified phase-1 construction, they are sampled from a
Gaussian conditioned on having norm at most $T_0$, with $T_0$
independent of the confidence parameter. The ideal corrector
$h_\star$ remains defined using the unconditioned Gaussian.

We first smooth $v$ to make it nearly
$1$-Lipschitz, then replace the expectation by a finite average.

\subsection{Smoothing the small oscillations}

Apply the Ornstein--Uhlenbeck operator at time $\beta$:
\[
 U=U_\beta(v),\qquad
 U(x)=\E_{\bz}v(ax+b\bz),\qquad
 a=e^{-\beta},\quad b=\sqrt{1-e^{-2\beta}},\quad \bz\sim\gauss.
\]

\begin{lemma}[Smoothing a uniformly small residual]
\label{lem:proper-smoothing}
If $\norm{v-h_\star}_\infty\le\beta$, then
\begin{equation}\label{eq:ideal}
 \Lip(U)\le1+\sqrt{\beta},\qquad
 \norm{U-v}_2\le\sqrt{\beta}+2\beta.
\end{equation}
\end{lemma}

\begin{proof}
The averaged witness $U_\beta(h_\star)$ has Lipschitz constant at
most $a$. To see this, couple the probes at $x$ and $y$. This bounds
the difference by $a\norm{x-y}$. For the residual $r$, differentiating
the Gaussian kernel gives, for every unit vector $u$,
\[
 \partial_u U_\beta(r)(x)
 =\frac ab\E_{\bz}[\langle u,\bz\rangle r(ax+b\bz)],\qquad
 |\partial_u U_\beta(r)(x)|
 \le\sqrt{\frac2\pi}\frac{a\beta}{b}\le\sqrt{\beta}.
\]
The last inequality uses $b\ge\sqrt\beta$ for $\beta\le1/16$.
Combining the two bounds gives
$\Lip(U)\le a+\sqrt{\beta}\le1+\sqrt{\beta}$.
By \Cref{fact:ou-smoothing-cost},
$\norm{U_\beta(h_\star)-h_\star}_2\le\sqrt{\beta}$.
Since the OU operator $U_\beta$ contracts $L_2(\gauss)$ and
$\norm r_2\le\beta$, we also have
$\norm{U-v}_2\le\sqrt{\beta}+2\beta$.
\end{proof}

\subsection{Approximating values and gradients by a finite average}

We must replace the expectation by a finite average for efficient
evaluation. To preserve the Lipschitz bound, we must control not
only the approximation of values but also that of gradients.

It suffices to obtain these bounds on a ball containing almost all
Gaussian mass. Set
\[
 R=\sqrt d+\sqrt{2\log(4L^2/\beta^2)},\qquad
 \mathcal B_R=\{x:\norm x\le R\}.
\]
Then $\Prb[\bx\notin\mathcal B_R]\le\beta^2/(4L^2)$.
Consequently, an error bounded by $2L$ outside the ball contributes
at most $\beta$ to the Gaussian $L_2$ error.

\paragraph{Structure of the empirical corrector.}
To show that a finite average approximates both values and gradients,
we use the structure of the empirical corrector $v$.
Let $s$ be the phase-1 smoothing time. At a query $y$, the empirical
corrector computes
\[
 u_i(y)=C_L(Q(\rho_s y+\sigma_s\bz_i)),\qquad i=1,\ldots,N,
\]
using the stored probes $\bz_1,\ldots,\bz_N$, sorts these
values as $u_{(1)}(y)\le\cdots\le u_{(N)}(y)$, and returns
\[
 v(y)=\min_{j\le N}\{u_{(j)}(y)+\psi(j/N)\}.
\]
The penalties $\psi(j/N)$ are constants independent of $y$.
Each clipped function $u_i$ has three possible pieces: the constants
$-L$ and $L$, and the polynomial $Q(\rho_s y+\sigma_s\bz_i)$.
Sorting and taking the minimum select one of these pieces and add
a fixed penalty. Thus $v$ is locally Lipschitz and piecewise polynomial.
Where the selected piece does not change,
\[
 \nabla v(y)=0
 \quad\text{or}\quad
 \nabla v(y)=\rho_s\nabla Q(\rho_s y+\sigma_s\bz_i)
 \quad\text{for some }i.
\]
Two distinct polynomial pieces can agree only where their difference
vanishes, a set of measure zero. If the pieces are identical, their
gradients agree as well. Therefore, choosing a piece by a fixed
ordering of the indices at ties gives the actual gradient of $v$
almost everywhere.

\paragraph{Bounds on sampled gradients.}
Throughout, the stored probes satisfy the supplied deterministic bound
\[
 \norm{\bz_i}\le T_0\qquad\text{for every }i\le N.
\]
The next lemma bounds how gradients grow with the smoothing probe's
norm and gives a bound on their second moment. These estimates
will control the sampling and truncation errors, respectively.

\begin{lemma}[Gradient growth of the empirical corrector]
\label{lem:proper-gradient-growth}
Suppose $\norm{\bz_i}\le T_0$ for every $i\le N$.
For a sufficiently large universal $C$, set
\[
 D=[C(d+k+1)(1+R+T_0)]^{C(k+1)}.
\]
Then, uniformly for $x\in\mathcal B_R$,
\begin{equation}\label{eq:derivative-growth}
 \norm{\nabla v(ax+bz)}\le D(1+\norm z)^k,
 \qquad
 \left(\E_{\bz}\norm{\nabla v(ax+b\bz)}^2\right)^{1/2}\le D.
\end{equation}
At ties, gradients use the fixed choice of polynomial piece above.
\end{lemma}

\begin{proof}
We now bound the gradient of $v$ at the points $ax+bz$ used by the
smoothing step. At such a point, each nonzero selected gradient has
the form
\[
 \rho_s\nabla Q\bigl(\rho_s(ax+bz)+\sigma_s\bz_i\bigr).
\]
Since $a,b,\rho_s,\sigma_s\le1$ and $x\in\mathcal B_R$, the argument
of $Q$ satisfies
\[
 \norm{\rho_s(ax+bz)+\sigma_s\bz_i}\le R+\norm z+T_0.
\]
Applying \Cref{fact:polynomial-growth} with $\norm Q_2<10$, and using
$1+R+T_0+\norm z\le(1+R+T_0)(1+\norm z)$, gives
\[
 \norm{\nabla v(ax+bz)}
 \le A_0(1+\norm z)^k,\qquad
 A_0=[C(d+k+1)(1+R+T_0)]^{k+1}.
\]
Thus the gradient grows at most polynomially with the smoothing
probe's norm. The Gaussian moment bound in the same fact gives
\[
 \left(\E_{\bz}\norm{\nabla v(ax+b\bz)}^2\right)^{1/2}
 \le A_0\left(\E_{\bz}(1+\norm\bz)^{2k}\right)^{1/2}
 \le A_0[C(d+k+1)]^{k/2}.
\]
The stated $D$ dominates both $A_0$ and
$A_0[C(d+k+1)]^{k/2}$, proving the two bounds.
\end{proof}

Since $\E(1+\norm\bz)^k<\infty$, this bound allows us to
differentiate inside the Gaussian expectation:
\[
 \nabla_x\E_{\bz}v(ax+b\bz)
 =a\,\E_{\bz}\nabla v(ax+b\bz).
\]

\paragraph{The finite average.}
We discard probes with unusually large norm so that all sampled
derivatives have a common bound. Set
\[
 T=\sqrt d+\sqrt{4\log\frac{8(L+D)}{\sqrt{\beta}}},\qquad
 \widetilde U(x)
 =\E_{\bz}[\one_{\{\norm\bz\le T\}}v(ax+b\bz)].
\]
Draw independent $\bw_1,\ldots,\bw_M\sim\gauss$, independently of the
stored probes $\bz_i$, and define
\begin{equation}\label{eq:A}
 A(x)=\frac1M\sum_{j=1}^M
 \one_{\{\norm{\bw_j}\le T\}}v(ax+b\bw_j).
\end{equation}
Every query uses the same probes. For each stored probe $\bw_j$,
the indicator $\one_{\{\norm{\bw_j}\le T\}}$ is constant as $x$ varies.
Thus each term in the average is either zero or the locally
Lipschitz function $v(ax+b\bw_j)$, so $A$ is locally Lipschitz.
Also, $|A|\le L$ because $|v|\le L$.

\begin{lemma}[Uniform approximation by a finite average]
\label{lem:proper-finite-average}
Fix stored probes satisfying $\norm{\bz_i}\le T_0$ for every $i\le N$,
and set
\[
 K=D(1+T)^k,\qquad V=Cd\log((k+2)(N+1)).
\]
If
\begin{equation}\label{eq:Mchoice}
 M\ge C\frac{d(L+K+1)^2}{\beta}
 \left[
 V\log\frac{Cd(L+K+1)}{\sqrt{\beta}}
 +\log\frac{C(d+1)}\delta
 \right],
\end{equation}
then, with probability at least $1-\delta/2$ over the probes $\bw_j$,
\begin{equation}\label{eq:finite-bounds}
 \sup_{x\in\mathcal B_R}|A(x)-U(x)|\le\sqrt{\beta},\qquad
 \norm{\nabla A(x)-\nabla U(x)}\le\sqrt{\beta}
 \quad\text{for a.e. }x\in\mathcal B_R.
\end{equation}
On this event,
$\Lip(A|_{\mathcal B_R})\le\Lip(U)+\sqrt{\beta}$.
Moreover, $M$ can be chosen to satisfy \eqref{eq:complexity}.
\end{lemma}

\begin{proof}[Proof overview]
We bound the truncation error and the sampling error separately.
For truncation, $|v|\le L$ controls the discarded values, while
the second-moment bound in \Cref{lem:proper-gradient-growth},
together with Cauchy--Schwarz, controls the discarded gradients.

To control sampling error, we use the pointwise bound in
\Cref{lem:proper-gradient-growth}, which says that each term
$v(ax+b\bw_j)$ with $\norm{\bw_j}\le T$ has gradient norm at most $K$.
The piecewise polynomial structure of $v$ gives pseudo-dimension
at most $V$ for the value and coordinate derivative classes.
Bounded uniform convergence then gives the sample bound
\eqref{eq:Mchoice}. Combining the truncation and sampling errors
proves \eqref{eq:finite-bounds}. Since $A$ is locally Lipschitz,
the gradient estimate also gives
$\Lip(A|_{\mathcal B_R})\le\Lip(U)+\sqrt{\beta}$.

The full proof, including the resource bound \eqref{eq:complexity},
is given in \Cref{app:approx-corrector-2}.
\end{proof}

\subsection{Making the Lipschitz bound exact}

\begin{algorithm}[H]
\caption{\textsc{SmoothCorrector}$(v,d,k,N,L,\beta,\delta,T_0)$}
\label{alg:phase-two}
\begin{algorithmic}[1]
\Require Query access to $v=G-c$ and the parameters of
\Cref{prop:proper}, including a bound $T_0\ge1$ on all stored
phase-1 probe norms.
\Ensure A representation of $v_{\mathrm{prop}}:\R^d\to[-L,L]$
using stored probes and the evaluation rule below; the caller restores $c$.
\Statex \textbf{Preprocessing (once).}

\State Set
\[
 a=e^{-\beta},\qquad b=\sqrt{1-e^{-2\beta}},\qquad
 R=\sqrt d+\sqrt{2\log(4L^2/\beta^2)}.
\]
\State Choose a sufficiently large universal $C$ and set
\[
 \begin{aligned}
 D&=[C(d+k+1)(1+R+T_0)]^{C(k+1)},\\
 T&=\sqrt d+\sqrt{4\log\frac{8(L+D)}{\sqrt{\beta}}},\\
 K&=D(1+T)^k,\qquad V=Cd\log((k+2)(N+1)).
 \end{aligned}
\]
\State Choose $M$ to be the ceiling of the right-hand side
of \eqref{eq:Mchoice}.
\State Draw and store independent $\bw_1,\ldots,\bw_M\sim\gauss$,
independently of the probes defining $v$.
Mark the probes satisfying $\norm{\bw_j}\le T$ for use in every query.

\Statex \textbf{Evaluation at a query $x$.}
\State Set $y=\proj_Rx$, the projection of $x$ onto $\mathcal B_R$.
\State \Return
\[
 v_{\mathrm{prop}}(x)
 =\frac{1}{M(1+2\sqrt{\beta})}
 \sum_{j:\,\norm{\bw_j}\le T}v(ay+b\bw_j).
\]
\Statex Normalize by the original sample size $M$, including discarded probes.
Every query reuses the stored probes; evaluation uses no fresh randomness.
\end{algorithmic}
\end{algorithm}

In \Cref{alg:phase-two}, the output is centered; the caller restores
$c$ after this correction. We now analyze the algorithm.

\begin{proof}[Proof of \Cref{prop:proper}]
Fix the stored phase-1 probes satisfying the supplied norm bound.
The preprocessing of \Cref{alg:phase-two} chooses $M$ as in
\Cref{lem:proper-finite-average}. Its evaluation rule forms $A$ using
\eqref{eq:A}, and returns
\begin{equation}\label{eq:output}
 v_{\mathrm{prop}}(x)=\frac{A(\proj_Rx)}{1+2\sqrt{\beta}}.
\end{equation}
All probes are stored during preprocessing. Each evaluation uses
at most $M$ queries to $v$, and $|v_{\mathrm{prop}}|\le L$.

By \Cref{lem:proper-finite-average}, with probability at least
$1-\delta/2$ over the new probes, the finite approximation bounds
hold. Fix such an outcome and suppose
$\norm{v-h_\star}_\infty\le\beta$.
\Cref{lem:proper-smoothing,lem:proper-finite-average} give
\[
 \Lip(U)\le1+\sqrt{\beta},\qquad
 \Lip(A|_{\mathcal B_R})\le1+2\sqrt{\beta}.
\]
Projection onto the ball is $1$-Lipschitz, so the rescaling gives
$\Lip(v_{\mathrm{prop}})\le1$.

On $\mathcal B_R$, projection leaves $x$ unchanged, so the error
comes from smoothing, sampling, and rescaling.
Outside the ball, both functions lie in $[-L,L]$, so the Gaussian
$L_2$ contribution is at most
$2L\sqrt{\Prb[\bx\notin\mathcal B_R]}\le\beta$.
Combining these bounds gives
\[
 \norm{v_{\mathrm{prop}}-v}_2
 \le
 \underbrace{\sqrt{\beta}+2\beta}_{\text{smoothing}}
 +\underbrace{\sqrt{\beta}}_{\text{sampling}}
 +\underbrace{2L\sqrt{\beta}}_{\text{rescaling}}
 +\underbrace{\beta}_{\text{outside }\mathcal B_R}
 \le8(L+1)\sqrt{\beta}.
\]
The success probability $1-\delta/2$ is at least the claimed
$1-\delta$.
\end{proof}

Finally, restoring the center gives
$G_{\mathrm{prop}}=c+v_{\mathrm{prop}}$.
On the same event, whenever $\norm{v-h_\star}_\infty\le\beta$,
this function is $1$-Lipschitz, takes values in $[c-L,c+L]$, and
satisfies
$\norm{G_{\mathrm{prop}}-G}_2\le8(L+1)\sqrt\beta$.
\section{Removing the label bound}
\label{sec:realizable-clipping}

The bounded-label learner will suffice even when the target has unbounded
range. We first estimate a center, then clip the centered labels. Since
clipping preserves Lipschitzness, the resulting learning problem is still
realizable. Gaussian concentration bounds the error introduced by clipping.

The centering lemma below only needs a variance bound, so it also applies
to the agnostic extension. In the realizable setting, Gaussian Poincar\'e
gives $\Var(\by)=\Var(g(\bx))\le\E\norm{\nabla g(\bx)}_2^2\le1$.

\begin{lemma}[Centering by a sample median]
\label{lem:unbounded-centering}
Let $\by$ have mean $c$ and variance at most $V^2$, where $V\ge1$,
and let $0<\delta<1/2$. Let $m$ be the median of an independent
odd preliminary sample of $n_0\ge8\log(2/\delta)$ labels. With probability
at least $1-\delta/2$, $|m-c|\le2V$. For each such fixed $m$, a fresh label
satisfies $\norm{\by-m}_2\le3V$.
\end{lemma}
\begin{proof}
By Chebyshev's inequality, $\Pr(|\by-c|>2V)\le1/4$. If the median is outside
$[c-2V,c+2V]$, at least half the preliminary labels are outside this
interval. Hoeffding's inequality bounds the probability of this event by
$e^{-n_0/8}\le\delta/2$.
On its complement, the triangle inequality gives
$\norm{\by-m}_2\le\norm{\by-c}_2+|c-m|\le3V$.
\end{proof}

We next bound the absolute error from clipping a Lipschitz function.
Write $C_T$ for clipping to $[-T,T]$, and take function norms under
Gaussian measure.

\begin{lemma}[Clipping in absolute distance]
\label{lem:l1-clipping}
Let $f:\R^d\to\R$ be $1$-Lipschitz with $\norm f_1\le3R$, where $R\ge1$.
For every $r>0$ and $T\ge6R+r$,
\[
 \norm{f-C_T(f)}_1\le\frac2r e^{-r^2/2}.
\]
\end{lemma}
\begin{proof}
Let $b$ be a median of $f(\bx)$. At least half the mass lies on the side
of $b$ away from zero, so $|b|\le2\E|f(\bx)|\le6R$.
Gaussian isoperimetry (\Cref{fact:isoperimetric}), applied to the sublevel
and superlevel sets at $b$, gives
$\Pr(|f(\bx)-b|>t)\le2e^{-t^2/2}$ for $t>0$. Consequently,
\[
 \norm{f-C_T(f)}_1
 =\int_T^\infty\Pr(|f(\bx)|>u)\,du
 \le2\int_r^\infty e^{-t^2/2}\,dt
 \le\frac2r e^{-r^2/2}.
\]
\end{proof}

\begin{corollary}[Clipping a realizable target]
\label{cor:realizable-clipping}
Suppose $\bx\sim\gauss$ and $\by=g(\bx)$ for a $1$-Lipschitz function
$g:\R^d\to\R$. Let $0<\eps\le1$ and $0<\delta<1/2$.
Take the sample median $m$ from \Cref{lem:unbounded-centering} and set
\[
 B=6+\sqrt{2\log(8/\eps)}.
\]
With probability at least $1-\delta/2$,
\[
 \norm{g-m-C_B(g-m)}_1\le\eps/4.
\]
For every fixed $m$, the labels $C_B(\by-m)$ are realized by the
$1$-Lipschitz function $C_B(g-m)$ with range in $[-B,B]$.
\end{corollary}
\begin{proof}
Gaussian Poincar\'e gives $\Var(\by)\le1$. Thus
\Cref{lem:unbounded-centering} with $V=1$ gives
$\norm{g-m}_1\le\norm{g-m}_2\le3$ with probability at least $1-\delta/2$.
For each such $m$, apply \Cref{lem:l1-clipping} with $R=1$ and
$r=\sqrt{2\log(8/\eps)}>1$. The resulting bound is
$2e^{-r^2/2}/r\le\eps/4$. The final assertion follows because
translation and clipping preserve Lipschitzness.
\end{proof}

\section{Main algorithm for realizable absolute loss}
\label{sec:main-proof}

We first combine the polynomial learner with the two correction stages
for bounded labels, then apply the clipping reduction from
\Cref{sec:realizable-clipping}. This gives the following guarantee.

\begin{theorem}[Realizable learning under absolute loss]
\label{thm:realizable}
Suppose $\bx\sim\gauss$ and $\by=g(\bx)$ almost surely for a $1$-Lipschitz
function $g:\R^d\to\R$. For $0<\eps\le1$ and $0<\delta<1/2$, there is
an algorithm returning a finite formula $H:\R^d\to\R$ such that, with
probability at least $1-\delta$,
\[
 \operatorname{Lip}(H)\le1,
 \qquad \err_1(H)=\norm{H-g}_1\le\eps.
\]
Its sample complexity, arithmetic training and evaluation costs, and
representation size are
\[
 (d+1)^{\poly(1/\eps)}\left(\log\frac2\delta\right)^3
\]
No label bound or knowledge of the label mean is required. Evaluation
uses no further randomness.
\end{theorem}

\begin{algorithm}[H]
\caption{\textsc{LearnLipschitz}$(\eps,\delta)$}
\label{alg:learn-lipschitz}
\begin{algorithmic}[1]
\Require Sample access to $(\bx,\by)$; $0<\eps\le1$; $0<\delta<1/2$.
\Ensure A finite formula for $H$, evaluated without fresh samples or randomness.
\Statex \textbf{Training.}

\State Draw an odd preliminary sample of
$n_0\ge8\log(2/\delta)$ labels and let $m$ be its median. Set
\[
 \begin{gathered}
 B=6+\sqrt{2\log(8/\eps)},\qquad \eta=\tau=\eps/4,\\
 r=\tau/64,\qquad \alpha=e^{-2/r^4}.
 \end{gathered}
\]
\State On fresh examples with labels $C_B(\by-m)$, run
\Cref{alg:learn-polynomial} with parameters $\eta,\alpha$
and failure probability $\delta/4$. Store its polynomial $p$
and let $k$ be its degree.
\State Read $c=\E[p]$ from the constant Hermite coefficient of $p$.
Set
\[
 L=r^{-1},\qquad
 \beta=\min\left\{r,\frac1{16},
 \left(\frac{\tau}{16(L+1)}\right)^2\right\},
 \qquad Q=U_{2r^2}(p-c).
\]
Compute $Q$ from the Hermite coefficients of $p$.

\State Set $\gamma=\delta/16$. Choose $N$ to be the probe count
for \Cref{alg:phase-one} with failure probability $\gamma^2/4$, and choose a deterministic
$A_0\ge1$, independent of $\gamma$, such that
$N\le A_0\log(2/\gamma)$. Set
\[
 T_0=\sqrt d+\sqrt{2\log(8A_0)}.
\]
Run \Cref{alg:phase-one} with supplied center $c$, accuracy $r$,
and uniform error $\beta$, replacing its probe count and sampling
step by the chosen $N$ and $N$ independent standard Gaussian
probes conditioned on having norm at most $T_0$.
Generate these probes by rejection sampling, allowing at most
$C(N+\log(2/\gamma))$ draws for a sufficiently large universal $C$.
If this limit is exceeded, return the constant function $H=0$.
Store the resulting $G=c+\widehat{\repair}[C_L(Q)]$.

\State Run \Cref{alg:phase-two} on $v=G-c$ with parameters
$L,\beta$, failure probability $\delta/8$, and the supplied norm
bound $T_0$ above.
Store the resulting formula $v_{\mathrm{prop}}$ and all probes.
\\
\Statex \textbf{Evaluation at a query $x$.}
\State \Return
\[
 H(x)=m+C_B\bigl(c+v_{\mathrm{prop}}(x)\bigr).
\]
\Statex The label median $m$ and polynomial center $c$ are stored separately.
Evaluation uses no fresh examples or randomness.
\end{algorithmic}
\end{algorithm}

The combined correction bound below controls the change in predictions
and will also be used for the extensions in the appendices.

\begin{proposition}[Finite Lipschitz correction]
\label{prop:finite-repair-interface}
Let $B\ge1$, $0<\tau\le1$, and $0<\xi<1/2$, and set $r=\tau/64$.
Given the coefficients of a degree-$k$ polynomial $p$ with
$\Gamma(p)\le\exp(-4/r^4)$, an algorithm returns a finite formula
$H:\R^d\to[-B,B]$ such that, with probability at least $1-\xi$,
\[
 \operatorname{Lip}(H)\le1,
 \qquad \norm{H-C_B(p)}_1\le\tau.
\]
Its arithmetic training and evaluation costs and representation size are
\[
 (d+1)^{\poly(k,1/\tau)}\poly(B)
 \left(\log\frac2\xi\right)^3.
\]
All randomness is fixed during training.
\end{proposition}

\begin{proof}
We compose \Cref{alg:phase-one,alg:phase-two}, as in the correction
steps of \Cref{alg:learn-lipschitz}, using accuracy parameter $r$.
Since the coefficients are available, we center at $c=\E p$,
the constant Hermite coefficient of $p$.
Gaussian Poincar\'e and the squared-defect bound give
\[
 \norm{p-c}_2\le\norm{\nabla p}_2
 \le1+\sqrt{\Gamma(p)}\le2.
\]
Choose
\[
 L=r^{-1},\qquad
 \beta=\min\left\{r,\frac1{16},
   \left(\frac{\tau}{16(L+1)}\right)^2\right\},
 \qquad \gamma=\xi/4.
\]
The VC bound in the proof of
\Cref{prop:uniform-corrector-approximation} allows a deterministic
phase-1 sample size, at failure probability $\gamma^2/4$, satisfying
\[
 N\le A_0\log(2/\gamma),\qquad
 A_0=(d+1)^{O(k+\tau^{-4})}\poly(1/\tau),
\]
where $A_0\ge1$ is chosen independently of $\gamma$.
Set
\[
 T_0=\sqrt d+\sqrt{2\log(8A_0)}.
\]
Draw the phase-1 probes independently from the standard Gaussian
conditioned on having norm at most $T_0$.

To justify this change, consider the original unconditioned probes
and let
\[
 E=\{\norm{\bz_i}\le T_0\text{ for every }i\le N\}.
\]
Gaussian concentration gives
$p_{\mathrm{out}}:=\Pr[\norm Z>T_0]\le1/(8A_0)$, so
\[
 \Pr(E)=(1-p_{\mathrm{out}})^N
 \ge e^{-2p_{\mathrm{out}}N}
 \ge(\gamma/2)^{1/4}.
\]
The modified joint distribution is the original distribution
conditioned on $E$. Consequently, if $F$ denotes failure of
the original phase-1 guarantee, then
\[
 \Pr(F\mid E)
 \le\frac{\Pr(F)}{\Pr(E)}
 \le\frac{\gamma^2}{4}(2/\gamma)^{1/4}
 \le\gamma.
\]
Generate the conditioned probes by rejection sampling, capped at
$C(N+\log(2/\gamma))$ draws. Since each draw is accepted with
probability at least $7/8$, a sufficiently large universal $C$
makes the probability of exceeding this cap at most $\gamma$.
On exceeding the cap, return the zero function.

Thus, except with probability at most $2\gamma=\xi/2$,
phase 1 produces a function $G$ with range
$[c-r^{-1},c+r^{-1}]$, with $\norm{G-p}_1\le31r$, and uniformly
$\beta$-close to the specified $1$-Lipschitz witness.

We next apply \Cref{prop:proper} in its supplied-$T_0$ form,
with clipping height $L=r^{-1}$ and failure probability $\xi/2$.
Its input polynomial is
$Q=U_{2r^2}(p-c)$, which has degree at most $k$ and
\[
 \norm Q_2\le\norm{p-c}_2\le2<10.
\]
Conditional on the stored phase-1 probes, the finite-average
guarantee fails with probability at most $\xi/2$.
On success, uniform $\beta$-closeness implies that the output
$G_{\mathrm{prop}}$ is $1$-Lipschitz and satisfies
\[
 \norm{G_{\mathrm{prop}}-G}_1
 \le\norm{G_{\mathrm{prop}}-G}_2
 \le8(L+1)\sqrt\beta\le\tau/2.
\]
Here $G_{\mathrm{prop}}=c+v_{\mathrm{prop}}$, with
$v_{\mathrm{prop}}$ returned by \Cref{alg:phase-two}.
Return $H=C_B(G_{\mathrm{prop}})$.
Clipping preserves Lipschitzness and contracts distances, so
\[
 \norm{H-C_B(p)}_1
 \le\norm{G_{\mathrm{prop}}-p}_1
 \le\tau/2+31r<\tau.
\]
A union bound gives failure probability at most $\xi$;
the success events need not be independent.

Finally, $\beta^{-1}$ is polynomial in $1/\tau$.
The quantities $T_0,D,T,K$ are independent of $\xi$ and satisfy
bounds of the form $(d+1)^{\poly(k,1/\tau)}$.
Writing $\ell=\log(2/\xi)$, we have $N\le CA_0\ell$ and
\[
 V=Cd\log((k+2)(N+1))
 \le Cd\bigl[
   \log(C(k+2)(A_0+1))+\log\ell
 \bigr].
\]
Consequently, \Cref{lem:finite-average-sample-size} permits
\[
 M\le(d+1)^{\poly(k,1/\tau)}\ell.
\]
Each evaluation uses $M$ evaluations of the empirical corrector,
each requiring $N$ polynomial evaluations and
$O(N\log(N+1))$ arithmetic operations for sorting.
Since
\[
 \log(N+1)\le\log(C(A_0+1))+\log\ell,
\]
the total evaluation cost is at most
\[
 (d+1)^{\poly(k,1/\tau)}\poly(B)\ell^3.
\]
Training and representation size satisfy the same upper bound.
The smoothed polynomial is computed from the Hermite coefficients
of $p$. Both stages store their probes and reuse them for every
query, so the output is a single finite formula.
\end{proof}

\begin{proposition}[Bounded-label realizable learning]
\label{thm:realizable-bounded}
Suppose $\bx\sim\gauss$ and $\by=g(\bx)$ almost surely, where
$g:\R^d\to[-B,B]$ is $1$-Lipschitz and $B\ge1$ is known.
For $0<\eps\le1$ and $0<\delta<1/2$, there is an algorithm returning
a finite formula $H:\R^d\to[-B,B]$ such that, with probability at least
$1-\delta$,
\[
 \operatorname{Lip}(H)\le1,
 \qquad \err_1(H)=\norm{H-g}_1\le\eps.
\]
Its sample complexity, arithmetic training and evaluation costs, and
representation size are
\[
 (d+1)^{\poly(B,1/\eps)}\left(\log\frac2\delta\right)^3.
\]
Evaluation uses no further randomness.
\end{proposition}

\begin{proof}
Set $\eta=\tau=\eps/2$, $r=\tau/64$, and
$\alpha=\exp(-2/r^4)$.
Run \Cref{alg:learn-polynomial}, analyzed in
\Cref{prop:polynomial-learning}, with parameters $\eta,\alpha$
and failure probability $\delta/2$.
On success, its output $p$ has $\err_1(p)\le\eps/2$ and
$\Gamma(p)\le\alpha^2=\exp(-4/r^4)$.
Apply \Cref{prop:finite-repair-interface} with accuracy $\tau$ and
failure probability $\delta/2$ to obtain $H$.
Clipping cannot increase loss against $\by\in[-B,B]$, so on the joint
success event
\[
 \err_1(H)\le \err_1(C_B(p))+\norm{H-C_B(p)}_1
 \le\eps/2+\tau=\eps.
\]
The correction also makes $H$ globally $1$-Lipschitz, and a union bound
gives success probability at least $1-\delta$.

The polynomial degree is
\[
 k=O\!\left(\frac{B^2}{\eps^2}
       \log\frac{B+1}{\alpha\eps}\right)
   =\poly(B,1/\eps).
\]
Although $1/\alpha$ is exponential in a polynomial in $1/\eps$, its
contribution to the polynomial learner's cost is absorbed by
$(d+1)^{\poly(B,1/\eps)}$.
By \Cref{lem:polynomial-confidence}, polynomial learning has at most
quadratic logarithmic dependence on $1/\delta$.
The correction has the same dimension and accuracy dependence,
with at most a cubic logarithmic dependence on $1/\delta$.
These costs add, giving the claimed resource bound.
\end{proof}

\begin{proof}[Proof of \Cref{thm:realizable}]
We analyze \Cref{alg:learn-lipschitz}.
Its polynomial and correction steps implement the bounded-label
learner of \Cref{thm:realizable-bounded} with accuracy $\eps/2$
and failure probability $\delta/2$.
Draw an odd preliminary sample of $n_0\ge8\log(2/\delta)$ labels
and let $m$ be its median.
Set $B=6+\sqrt{2\log(8/\eps)}$.
On fresh samples with labels $C_B(\by-m)$, apply
\Cref{thm:realizable-bounded} with accuracy $\eps/2$ and failure
probability $\delta/2$, obtaining $\widehat h:\R^d\to[-B,B]$.
If the rejection-sampling cap is exceeded, the algorithm instead returns
$H=0$; this event is already included in the bounded-label learner's
failure budget. Otherwise, return $H=m+\widehat h$.

By \Cref{cor:realizable-clipping}, except with probability $\delta/2$,
$\norm{g-m-C_B(g-m)}_1\le\eps/4$.
For every fixed median, the clipped problem is realized by
$C_B(g-m)$, so the bounded-label learner succeeds with conditional
probability at least $1-\delta/2$.
On the joint success event, adding $m$ preserves Lipschitzness and
\[
 \err_1(H)=\norm{m+\widehat h-g}_1
 \le\norm{\widehat h-C_B(g-m)}_1
     +\norm{C_B(g-m)-(g-m)}_1
 \le\eps/2+\eps/4\le\eps.
\]
A union bound gives success probability at least $1-\delta$.

Since $B=O(\sqrt{\log(8/\eps)})$, substituting this bound and accuracy
$\eps/2$ into \Cref{thm:realizable-bounded} gives the claimed resources.
The preliminary sample costs $O(\log(1/\delta))$ additional examples,
and evaluation only adds the stored center $m$ to $\widehat h$.
\end{proof}

\appendix
\section{Proofs for learning a polynomial with small squared defect}
\label[appendix]{sec:appendix-learning-polynomial}

\subsection{Uniform convergence}
\label{sec:appendix-uniform-convergence}

Here, we give a proof of \Cref{prop:uniform}
using well-understood standard techniques in statistical learning theory;
see~\cite[Sections~4--5]{BousquetBoucheronLugosi2004} for an overview of these techniques.

\paragraph{Facts about polynomials under Gaussian distribution.}

We use the normalized Hermite polynomials as an orthonormal basis for polynomials
with respect to the standard Gaussian measure.
The following fact collects the properties we need;
see~\cite[Chapter~11]{ODonnell2021} for background.

\begin{fact}[Hermite polynomials {\cite[Chapter~11]{ODonnell2021}}]
\label{fact:hermite-uniform}
For every multi-index $\nu \in \mathbb Z_{\ge0}^d$ there is a polynomial $H_\nu$ on $\R^d$,
the normalized Hermite polynomial, such that the following hold.
\begin{enumerate}
  \item[\textup{(i)}]
    $H_\nu$ has degree $|\nu| = \sum_{j=1}^d \nu_j$, and for every integer $m \ge 0$
    the family $\{H_\nu : |\nu| \le m\}$ is an orthonormal basis,
    with respect to the inner product
    $\langle f,g\rangle = \E_{\bx\sim\gauss}[f(\bx)g(\bx)]$,
    of the space of polynomials of degree at most $m$.
    In particular, this space has dimension $\binom{d+m}{m}$.
  \item[\textup{(ii)}]
    For every $j \in \{1,\ldots,d\}$,
    $\partial_j H_\nu = \sqrt{\nu_j} H_{\nu-e_j}$,
    where $e_j$ is the $j$-th standard basis vector
    and the right-hand side is $0$ when $\nu_j = 0$.
\end{enumerate}
\end{fact}

We also use the following standard consequence of Gaussian hypercontractivity,
which turns a first-moment bound on a polynomial into a second-moment bound.

\begin{fact}[Gaussian hypercontractivity corollary
{\cite[Proposition B.22]{CKKSV2024}}]
\label{fact:hypercontractivity-uniform}
Every polynomial $q:\R^d\to\R$ of degree at most $r$ satisfies
\[
  \left(\E_{\bx\sim\gauss}[q(\bx)^2]\right)^{1/2}
  \le e^r \E_{\bx\sim\gauss}[|q(\bx)|].
\]
\end{fact}

\paragraph{Rademacher complexity.}

For a class $\mathcal F$ of real-valued functions on $\R^d$,
define its Rademacher complexity by
\[
  \operatorname{Rad}_N(\mathcal F)
  := \E_{\bs,\bsigma}\left[
       \sup_{f\in\mathcal F}
       \left| \frac1N \sum_{i=1}^N \bsigma_i f(\bx_i) \right|
     \right],
\]
where $\bs=(\bx_1,\ldots,\bx_N)$ consists of independent standard Gaussians and the $\bsigma_i$ are
independent uniform signs in $\{-1,1\}$, independent of the sample.
Rademacher complexity is useful because a bound on it gives uniform convergence
through \Cref{fact:symmetrization-uniform} below,
and \Cref{fact:contraction-uniform} controls how it changes when the same Lipschitz function
is applied to every member of a class.
We state both facts only for classes of continuous functions, as this is all we need here;
this also ensures that all suprema involved are measurable.%
\footnote{For continuous $f$, the quantities inside the suprema below are continuous
  functions of the sample and the signs.
  A supremum of continuous functions is lower semicontinuous and hence Borel measurable,
  so every supremum below is a measurable random variable.
  See also Kanade~\cite[Section~4]{Kanade2018} for an elementary discussion of
  Rademacher complexity and contraction for bounded classes.}

\begin{fact}[Rademacher Complexity Implies Uniform Convergence
{\cite[Lemma 6.3]{LedouxTalagrand1991}}]
\label{fact:symmetrization-uniform}
Let $\mathcal F$ be a nonempty class of continuous functions on $\R^d$,
each integrable with respect to $\gauss$. Then
\[
  \E_{\bs} \left[
    \sup_{f\in\mathcal F}
    \bigg| \E_{\bx\in \bs}[f(\bx)] - \E_{\bx\sim\gauss}[f(\bx)] \bigg|
  \right]
  \le 2\operatorname{Rad}_N(\mathcal F).
\]
\end{fact}

\begin{fact}[Rademacher contraction {\cite[Theorem 4.12]{LedouxTalagrand1991}}]
\label{fact:contraction-uniform}
Let $\mathcal F$ be a nonempty class of nonnegative continuous functions on $\R^d$.
Let $\phi:[0,\infty)\to\R$ be $1$-Lipschitz with $\phi(0)=0$,
and define the composed class
$\phi\circ\mathcal F := \{\phi\circ f : f\in\mathcal F\}$. Then
\[
  \operatorname{Rad}_N(\phi\circ\mathcal F)
  \le 2\operatorname{Rad}_N(\mathcal F).
\]
\end{fact}

\begin{proof}[Proof of \Cref{prop:uniform}]
When $k=0$, every defect is zero. Assume henceforth that $k\ge1$.
In order to prove \Cref{prop:uniform}, the class for which we want to show
uniform convergence is
\[
  \mathcal F_{\mathrm{grad\text{-}defect},k}
  := \left\{
       x \mapsto (\norm{\nabla P(x)}-1)_+^2 :
       \deg P \le k,\ \E_{\bx\sim\gauss}[P(\bx)^2] \le A
     \right\}.
\]
By \Cref{fact:symmetrization-uniform}, to show \Cref{prop:uniform} it is enough to bound
the Rademacher complexity of $\mathcal F_{\mathrm{grad\text{-}defect},k}$.
We do this by using the contraction lemma, \Cref{fact:contraction-uniform}.
To apply it, first define the class of squared gradient norms
\[
  \mathcal F_{\mathrm{grad\text{-}norm},k}
  := \left\{
       x \mapsto \norm{\nabla P(x)}^2 :
       \deg P \le k,\ \E_{\bx\sim\gauss}[P(\bx)^2] \le A
     \right\}.
\]
The function $\phi(t) = (\sqrt t-1)_+^2$ is zero on $[0,1]$ and has derivative
$1-1/\sqrt t \in [0,1]$ for $t>1$.
Thus it is $1$-Lipschitz, $\phi(0)=0$, and
$\mathcal F_{\mathrm{grad\text{-}defect},k}
=\phi\circ\mathcal F_{\mathrm{grad\text{-}norm},k}$,
which allows us to use
\Cref{fact:symmetrization-uniform,fact:contraction-uniform} to conclude that
\begin{multline}\label{eq:uniform-reduction}
  \E_{\bs} \left[
    \sup_{f\in\mathcal F_{\mathrm{grad\text{-}defect},k}}
    \bigg| \E_{\bx\in \bs}[f(\bx)] - \E_{\bx\sim\gauss}[f(\bx)] \bigg|
  \right] \\
  \le 2\operatorname{Rad}_N(\mathcal F_{\mathrm{grad\text{-}defect},k})
  \le 4\operatorname{Rad}_N(\mathcal F_{\mathrm{grad\text{-}norm},k}).
\end{multline}
Consequently, it remains to bound the Rademacher complexity of
$\mathcal F_{\mathrm{grad\text{-}norm},k}$.

From now on, we focus on obtaining this bound.
First, every polynomial $Q(x) = \norm{\nabla P(x)}^2$ in this class satisfies
$\E_{\bx\sim\gauss}[Q(\bx)^2] \le k^2A^2e^{4k-4}$.
To see why, expand $P = \sum_{|\nu|\le k} a_\nu H_\nu$ in the normalized Hermite basis
of \Cref{fact:hermite-uniform}.
Orthonormality of Hermite polynomials gives
$\sum_\nu a_\nu^2 = \E_{\bx\sim\gauss}[P(\bx)^2] \le A$.
Using the non-negativity of $\norm{\nabla P(x)}^2$,
the derivative identity $\partial_j H_\nu = \sqrt{\nu_j} H_{\nu-e_j}$
from \Cref{fact:hermite-uniform},
and orthonormality of Hermite polynomials once more, we get
\begin{equation}\label{eq:uniform-first-moment}
  \E_{\bx\sim\gauss}[|Q(\bx)|]
  = \E_{\bx\sim\gauss}[\norm{\nabla P(\bx)}^2]
  = \sum_{|\nu|\le k} |\nu| a_\nu^2
  \le kA.
\end{equation}
Since $\deg Q \le 2k-2$, \Cref{fact:hypercontractivity-uniform}
applied to \Cref{eq:uniform-first-moment} gives
\[
  \E_{\bx\sim\gauss}[Q(\bx)^2]
  \le e^{4k-4} \bigl(\E_{\bx\sim\gauss}[|Q(\bx)|]\bigr)^2
  \le k^2A^2e^{4k-4}
  =: R^2.
\]
Thus every function $f$ in $\mathcal F_{\mathrm{grad\text{-}norm},k}$
is a polynomial of degree at most $2k-2$
with $\E_{\bx\sim\gauss}[f(\bx)^2] \le R^2$.
We now show that this is enough to upper-bound the Rademacher complexity of
$\mathcal F_{\mathrm{grad\text{-}norm},k}$.
Let $h_1,\ldots,h_M$ be the normalized Hermite polynomials of degree at most $2k-2$,
which by \Cref{fact:hermite-uniform} form an orthonormal basis for polynomials of degree
at most $2k-2$, where $M = \binom{d+2k-2}{2k-2}$.
Writing $f = \sum_{j=1}^M b_j h_j$, we have $\sum_j b_j^2 \le R^2$.
For every fixed sample and choice of signs, Cauchy--Schwarz therefore gives
\[
\begin{aligned}
  \sup_{f\in\mathcal F_{\mathrm{grad\text{-}norm},k}}
  \left| \sum_{i=1}^N \bsigma_i f(\bx_i) \right|
  &\le \sup_{\substack{\deg f \le 2k-2 \\
                       \E_{\bx\sim\gauss}[f(\bx)^2] \le R^2}}
       \left| \sum_{i=1}^N \bsigma_i f(\bx_i) \right| \\
  &\le \sup_{\sum_j b_j^2 \le R^2}
       \left| \sum_{j=1}^M b_j \sum_{i=1}^N \bsigma_i h_j(\bx_i) \right| \\
  &\le R \left(
         \sum_{j=1}^M \left( \sum_{i=1}^N \bsigma_i h_j(\bx_i) \right)^2
       \right)^{1/2}.
\end{aligned}
\]
The first inequality enlarges the class $\mathcal F_{\mathrm{grad\text{-}norm},k}$
to all polynomials of degree at most $2k-2$
with $\E_{\bx\sim\gauss}[f(\bx)^2] \le R^2$;
taking a larger class can only make the supremum larger.
The second inequality expands each such $f$ in the Hermite basis $h_1,\ldots,h_M$.
Substituting the inequality above into the definition of Rademacher complexity,
and using Jensen's inequality yields
\begin{equation}\label{eq:uniform-complexity}
\begin{aligned}
  \operatorname{Rad}_N(\mathcal F_{\mathrm{grad\text{-}norm},k})
  &= \E_{\bs,\bsigma}\left[
       \sup_{f\in\mathcal F_{\mathrm{grad\text{-}norm},k}}
       \left| \frac1N \sum_{i=1}^N \bsigma_i f(\bx_i) \right|
     \right] \\
  &\le \frac RN \E_{\bs,\bsigma}\left[
        \left(
          \sum_{j=1}^M \left( \sum_{i=1}^N \bsigma_i h_j(\bx_i) \right)^2
        \right)^{1/2}
      \right] \\
  &\le \frac RN \left(
        \sum_{j=1}^M \E_{\bs,\bsigma}\left[
          \left( \sum_{i=1}^N \bsigma_i h_j(\bx_i) \right)^2
        \right]
      \right)^{1/2}
  = R\sqrt{\frac MN}.
\end{aligned}
\end{equation}
The last step above follows because each expectation inside the sum is $N$:
after expanding the square, the independent signs cancel the cross terms,
and orthonormality of the Hermite polynomials $\{h_j\}$ gives
$\E_{\bx\sim\gauss}[h_j(\bx)^2] = 1$.

It remains to substitute $M$, $R$ and $N$ into the inequality above
and check that \Cref{prop:uniform} indeed follows.
Set $W = A(d+1)^k$ and $u = \beta\delta \le 1/4$,
so that the assumption on $N$ in \Cref{prop:uniform} reads $N \ge (W/u)^{10}$.
Recalling that $M = \binom{d+2k-2}{2k-2}$, we have
$\sqrt M \le (d+1)^{k-1} \le W$ and
$R = Ake^{2k-2} \le Ak9^{k-1} \le A10^k \le W^4$,
where $k9^{k-1} \le 10^k$ follows from the binomial expansion of $(9+1)^k$.
\Cref{eq:uniform-reduction,eq:uniform-complexity} give
\[
  \E_{\bs} \left[
    \sup_{f\in\mathcal F_{\mathrm{grad\text{-}defect},k}}
    \bigg| \E_{\bx\in \bs}[f(\bx)] - \E_{\bx\sim\gauss}[f(\bx)] \bigg|
  \right]
  \le \frac{4W^5}{\sqrt N}
  \le 4u^5
  \le u
  = \beta\delta,
\]
where the second inequality uses the assumption $N \ge (W/u)^{10}$
and the third uses $u \le 1/4$.
By Markov's inequality, with probability at least $1-\delta$ over $\bs$,
\[
  \sup_{f\in\mathcal F_{\mathrm{grad\text{-}defect},k}}
  \bigg| \E_{\bx\in \bs}[f(\bx)] - \E_{\bx\sim\gauss}[f(\bx)] \bigg|
  \le \beta,
\]
which is the desired simultaneous uniform convergence.
This completes the proof of \Cref{prop:uniform}.
\end{proof}

We also need uniform convergence of the prediction loss.

\begin{lemma}[Uniform convergence of polynomial loss]
\label{lem:uniform-polynomial-loss}
Let $\bs$ consist of $n$ independent labeled examples.
For every $0<\beta,\delta\le1/2$, if
\[
  n\ge
  \left(\frac{(B+1)(d+1)^k}{\beta\delta}\right)^{10},
\]
then with probability at least $1-\delta$, every polynomial
$p$ of degree at most $k$ satisfying $\norm p_2\le B$ obeys
\[
  |\widehat{\err}_\ell(p)-\err_\ell(p)|\le\beta.
\]
\end{lemma}

\begin{proof}
Let $\mathcal P$ be the class of polynomials of degree at
most $k$ with $\norm p_2\le B$, and put
$D=\binom{d+k}{k}$. The Hermite basis gives
$\operatorname{Rad}_n(\mathcal P)\le B\sqrt{D/n}$.
For absolute loss, symmetrization and contraction applied to
$t\mapsto|t-y|-|y|$ give
\[
  \E_{\bs} \sup_{p\in\mathcal P}
  |\widehat{\err}_1(p)-\err_1(p)|
  \le\frac{4B\sqrt D+B}{\sqrt n}.
\]
For squared loss, expand $(p-y)^2=p^2-2yp+y^2$.
Hypercontractivity gives
$\norm{p^2}_2=\norm p_4^2\le3^kB^2$.
The Hermite basis and symmetrization therefore give
\[
  \E_{\bs} \sup_{p\in\mathcal P}
  |\widehat{\err}_2(p)-\err_2(p)|
  \le
  \frac{B^2}{\sqrt n}
  \left(
    2\cdot3^k\sqrt{\binom{d+2k}{2k}}
    +4\sqrt D+1
  \right).
\]
Since $\binom{d+2k}{2k}\le D^2$, both bounds imply
\begin{equation}\label{eq:uniform-loss-expectation}
  \E_{\bs} \sup_{p\in\mathcal P}
  |\widehat{\err}_\ell(p)-\err_\ell(p)|
  \le\frac{7B^2 3^kD}{\sqrt n}.
\end{equation}
Using $D\le(d+1)^k$, the assumed sample size makes this
at most $\beta\delta$. Markov's inequality completes the proof.
\end{proof}

\subsection{Empirical polynomial learning}

Write polynomials of degree at most $k$ as
\[
  p_c(x)=c^T\bH(x),
  \qquad D=\binom{d+k}{k},
\]
in the normalized Hermite basis, so $\norm{p_c}_2=\norm c$.
Let $\bA(x)$ have columns $\nabla H_\nu(x)$, so
$\nabla p_c(x)=\bA(x)c$. For a labeled sample
$\bs=\{(\bx_i,\by_i)\}_{i=1}^n$, define
\[
  \widehat{\err}_\ell(p)=\frac1n\sum_i|p(\bx_i)-\by_i|^\ell,
  \qquad
  \widehat\Gamma(p)=\frac1n\sum_i
  (\norm{\nabla p(\bx_i)}-1)_+^2.
\]

For $0<\eta,\alpha\le1$, set
\[
  b=\frac{\eta}{4(4B+1)},
  \qquad t=\frac{b^2}{4},
\]
and choose
\begin{equation}\label{eq:degree-choice}
  k=\max\left\{
    1,\left\lceil
    t^{-1}\log\max\left\{\frac2b,\frac2\alpha\right\}
    \right\rceil
  \right\}.
\end{equation}

\begin{proposition}[Polynomial learning for bounded agnostic labels]
\label{prop:polynomial-learning-agnostic}
Suppose $\bx\sim\gauss$, $|\by|\le B$, and $B\ge1$. For either
$\ell\in\{1,2\}$ and $0<\eta,\alpha,\zeta\le1$, there is an algorithm
that, with probability at least $1-\zeta$, returns a polynomial $p$ with
\[
 \err_\ell(p)\le\OPT_\ell+\eta,\qquad
 \Gamma(p)\le\alpha^2,\qquad \norm p_2\le B.
\]
Its degree is given by \Cref{eq:degree-choice} with parameters
$\eta/2,\alpha/2$, and its resource bounds
are those in \Cref{prop:polynomial-learning}.
\end{proposition}
This implies \Cref{prop:polynomial-learning} by taking $\ell=1$ and
$\OPT_1=0$.

\begin{proof}[Proof of \Cref{prop:polynomial-learning-agnostic}]
We first give the base sample implementation of \Cref{alg:learn-polynomial}
for $\ell=1$ and its squared-loss extension for $\ell=2$.
\Cref{lem:polynomial-confidence} then applies this implementation
at constant confidence with tighter accuracy and defect parameters
to obtain the stated resource bounds.

Choose a bounded $1$-Lipschitz comparator $g$ with
$\err_\ell(g)\le\OPT_\ell+\eta/4$.
By \Cref{lem:low-degree-witness} and the choice of $t,k$, the polynomial
$Q=(U_tg)^{\le k}$ has $\norm{Q-g}_2\le b$ and
$\Gamma(Q)\le\alpha^2/4$. Its absolute loss exceeds that of $g$
by at most $b$. For squared loss, Cauchy--Schwarz gives
\[
 |\err_2(Q)-\err_2(g)|
 \le\norm{Q-g}_2\bigl(\norm{Q-\by}_2+\norm{g-\by}_2\bigr)
 \le4Bb,
\]
since $\norm Q_2,\norm g_2,\norm \by_2\le B$.
Our choice of $b$ therefore gives
\[
  \norm Q_2\le B,\qquad
  \err_\ell(Q)\le\OPT_\ell+\eta/2,\qquad
  \Gamma(Q)\le\alpha^2/4.
\]

To implement the population program, minimize over $\norm c\le B$
the empirical objective
\begin{equation}\label{eq:empirical-shifted-objective}
 \widehat F(c)
 =\widehat{\err}_\ell(p_c)
 +M\left(\widehat\Gamma(p_c)-\frac{\alpha^2}{2}\right)_+,
 \qquad M=\frac{4(B^\ell+1)}{\alpha^2}.
\end{equation}
This objective is convex: $(\norm v-1)_+^2$ is the squared distance
to the Euclidean unit ball, and the positive part preserves convexity.
The penalty allows us to optimize over the coefficient ball alone.
We prove the required population guarantees directly, allowing the
reserved slack between $\alpha^2/2$ and $\alpha^2$.

Take
\[
  n=\left\lceil
    \frac{CB^4D^2}{\zeta^2}
    \left(
      \frac{k^2e^{4k}}{\alpha^4}
      +\frac{9^k}{\eta^2}
    \right)
  \right\rceil
\]
for a sufficiently large universal constant $C$.
Using $\binom{d+2k}{2k}\le D^2$, the bounds
\Cref{eq:uniform-reduction,eq:uniform-complexity,eq:uniform-loss-expectation} and Markov's inequality imply
that, with probability at least $1-\zeta/2$, simultaneously
for all $\deg p\le k$ with $\norm p_2\le B$,
\begin{equation}\label{eq:empirical-uniform-event}
  |\widehat{\err}_\ell(p)-\err_\ell(p)|\le\eta/8,
  \qquad
  |\widehat\Gamma(p)-\Gamma(p)|\le\alpha^2/8.
\end{equation}
On this event, $\widehat\Gamma(Q)\le3\alpha^2/8$, so the witness
incurs no penalty: $\widehat F(c_Q)=\widehat{\err}_\ell(Q)$.
Suppose $\norm{\bar c}\le B$ and
$\widehat F(\bar c)\le\min_{\norm c\le B}\widehat F(c)+\eta/8$.
Comparison with $Q$ gives
\[
 \err_\ell(p_{\bar c})
 \le\widehat F(\bar c)+\eta/8
 \le\err_\ell(Q)+3\eta/8
 \le\OPT_\ell+7\eta/8.
\]
The zero polynomial has objective at most $B^\ell$. Since the loss
and penalty are nonnegative,
\[
 \widehat\Gamma(p_{\bar c})
 \le\frac{\alpha^2}{2}+\frac{\widehat F(\bar c)}{M}
 \le\frac{\alpha^2}{2}+\frac{B^\ell+\eta/8}{M}
 \le\frac{3\alpha^2}{4}.
\]
Uniform convergence now gives
$\Gamma(p_{\bar c})\le7\alpha^2/8\le\alpha^2$.

To find this approximate minimizer, we use projected subgradient descent
on the radius-$B$ coefficient ball. Loss subgradients are obtained by
averaging those of $|p_c(\bx_i)-\by_i|^\ell$. The empirical defect has gradient
\[
 \nabla_c\widehat\Gamma(p_c)
 =\frac2n\sum_{i=1}^n\bA(\bx_i)^T
 \bigl(\bA(\bx_i)c-\proj_1(\bA(\bx_i)c)\bigr),
\]
where $\proj_1$ denotes projection onto the Euclidean unit ball.
Multiply this gradient by $M$ when the defect exceeds $\alpha^2/2$,
and by zero otherwise; either choice is valid at equality.

We bound the subgradient norms using second moments of the features.
Orthonormality and the Hermite derivative identity give
\[
 \E\left[\frac1n\sum_i\norm{\bH(\bx_i)}^2\right]=D,
 \qquad
 \E\left[\frac1n\sum_i\norm{\bA(\bx_i)}_F^2\right]\le kD.
\]
Thus, with probability at least $1-\zeta/2$,
\[
 \frac1n\sum_i\norm{\bH(\bx_i)}^2\le H_0^2:=\frac{4D}{\zeta},
 \qquad
 \frac1n\sum_i\norm{\bA(\bx_i)}_F^2\le A_0^2:=\frac{4kD}{\zeta}.
\]
On this event, loss subgradients have norm at most $H_0$ for absolute
loss and $2B(H_0^2+H_0)$ for squared loss. Also,
$\norm{v-\proj_1v}\le\norm v$ bounds the penalty subgradient by
$2MBA_0^2$. Hence every subgradient of $\widehat F$ on the ball has
norm at most
\[
 G_0=2(B+1)(H_0^2+H_0)+2MBA_0^2.
\]

Initialize at zero and run projected subgradient descent for
\[
 T=\left\lceil\left(\frac{16BG_0}{\eta}\right)^2\right\rceil
 \qquad\text{iterations with step size}\qquad
 a=\frac{2B}{G_0\sqrt T}.
\]
Let $c_\star$ minimize $\widehat F$ on the ball and let $g_j$ be a
subgradient at the iterate $c_j$. Projection does not increase the
distance to $c_\star$, so
\[
 \norm{c_{j+1}-c_\star}^2
 \le\norm{c_j-c_\star}^2
 -2a\langle g_j,c_j-c_\star\rangle+a^2G_0^2.
\]
Summing telescopes the squared distances. The subgradient inequality
and convexity then give, for the average $\bar c$ of the $T$ iterates,
\[
 \widehat F(\bar c)-\widehat F(c_\star)
 \le\frac{(2B)^2}{2aT}+\frac{aG_0^2}{2}
 =\frac{2BG_0}{\sqrt T}\le\frac\eta8.
\]
The average remains in the coefficient ball, so $\norm{p_{\bar c}}_2\le B$.

Each iteration takes polynomial time in $n,D,d,k$. Since
$e^{4k}\le(9^k)^2$, both $n$ and $T$ are polynomial in
$B,\eta^{-1},\alpha^{-1},\zeta^{-1},k,D,9^k$. The algorithm uses the same fixed iteration budget on every
sample. A union bound over the feature-moment event and
\eqref{eq:empirical-uniform-event} completes the proof.
\end{proof}

\subsection{Confidence amplification for polynomial learning}
\label{sec:polynomial-confidence}

The preceding empirical implementation has polynomial dependence on
 the inverse failure probability. We improve this dependence by running
it at constant confidence and independently validating both the loss
and the squared defect.

\begin{lemma}[Confidence amplification]
\label{lem:polynomial-confidence}
Under the assumptions of
\Cref{prop:polynomial-learning-agnostic}, there is an algorithm
satisfying its guarantees with sample complexity and arithmetic
running time at most
\[
 P\left(B,\eta^{-1},\alpha^{-1},k,D,9^k\right)
 \left(\log\frac2\zeta\right)^2,
\]
where $P$ is a universal polynomial,
$D=\binom{d+k}{k}$, and $k$ is chosen using
\eqref{eq:degree-choice} with parameters $\eta/2,\alpha/2$.
\end{lemma}

\begin{proof}
It suffices to consider $0<\zeta<1/2$; for larger $\zeta$ use
confidence $1/4$.
Run the preceding base empirical implementation independently
\[
 J=\left\lceil\log_4\frac4\zeta\right\rceil
\]
times, with accuracy $\eta/2$, defect parameter $\alpha/2$,
and failure probability $1/4$.
Let $p_1,\ldots,p_J$ be the resulting polynomials.
Every output satisfies $\norm{p_j}_2\le B$, regardless of whether
the run succeeds, because the implementation projects onto the
radius-$B$ coefficient ball.
With probability at least $1-\zeta/4$, at least one candidate satisfies
\[
 \err_\ell(p_j)\le\OPT_\ell+\eta/2,
 \qquad \Gamma(p_j)\le\alpha^2/4.
\]

We estimate the loss and squared defect of all candidates using
 a fresh common validation sample.
For every degree-$\le k$ polynomial $p$ with $\norm p_2\le B$,
Gaussian hypercontractivity gives
\[
 \E p(\bx)^4\le9^kB^4.
\]
Also, applying hypercontractivity to each partial derivative and
then using Cauchy--Schwarz gives
\[
 \begin{aligned}
 \E\norm{\nabla p(\bx)}^4
 &=\sum_{i,h}
 \E\bigl[(\partial_i p(\bx))^2(\partial_h p(\bx))^2\bigr]\\
 &\le\left(\sum_i\norm{\partial_i p}_4^2\right)^2\\
 &\le9^{k-1}\left(\sum_i\norm{\partial_i p}_2^2\right)^2
 \le9^{k-1}k^2B^4.
 \end{aligned}
\]
Since $|\by|\le B$, both validation variables
\[
 |p(\bx)-\by|^\ell,
 \qquad (\norm{\nabla p(\bx)}-1)_+^2
\]
have variance at most
\[
 W=C9^k(k+1)^2(B+1)^4,
 \qquad \ell\in\{1,2\}.
\]

Partition the validation sample into an odd number
$q=O(\log(8J/\zeta))$ of independent blocks, with
$q\ge C\log(8J/\zeta)$, each containing
\[
 s=\left\lceil CW(\eta^{-2}+\alpha^{-4})\right\rceil
\]
examples.
For each candidate, estimate its loss and squared defect by the
median of the corresponding block averages.
Write these estimates as $\widetilde e_j$ and
$\widetilde\Gamma_j$.
Conditional on the training samples, the candidates are fixed.
Chebyshev's inequality makes the failure probability of each
block estimate at most $1/16$, after increasing $C$.
A binomial tail bound for the median and a union bound over the
$2J$ estimates therefore give, with probability at least $1-\zeta/2$,
\[
 |\widetilde e_j-\err_\ell(p_j)|\le\eta/8,
 \qquad
 |\widetilde\Gamma_j-\Gamma(p_j)|\le\alpha^2/8
 \quad\text{for every }j\le J.
\]

Retain the candidates satisfying
$\widetilde\Gamma_j\le\alpha^2/2$, and return one with the
smallest estimated loss. If none is retained, return the zero polynomial.
On the joint success event, a successful base candidate is retained,
since
\[
 \widetilde\Gamma_j\le\alpha^2/4+\alpha^2/8=3\alpha^2/8.
\]
The selected candidate $p$ satisfies
\[
 \Gamma(p)\le\alpha^2/2+\alpha^2/8=5\alpha^2/8\le\alpha^2.
\]
Comparing its estimated loss with that of a successful candidate
$p_j$ gives
\[
 \err_\ell(p)
 \le\widetilde e_p+\eta/8
 \le\widetilde e_j+\eta/8
 \le\err_\ell(p_j)+\eta/4
 \le\OPT_\ell+3\eta/4.
\]
Its coefficient norm is at most $B$.
The total failure probability is at most $\zeta/4+\zeta/2\le\zeta$.

There are $J=O(\log(2/\zeta))$ constant-confidence training runs.
The validation sample has
$qs=\poly(B,\eta^{-1},\alpha^{-1},k,9^k)\log(2/\zeta)$
examples. Evaluating all $J$ candidates on it costs a polynomial
in the stated parameters times $(\log(2/\zeta))^2$.
This proves the resource bound.
\end{proof}

\section{Uniform approximation by a finite average for the smoothing post-processing}
\label[appendix]{app:approx-corrector-2}

We use standard techniques from statistical learning theory to prove
the uniform convergence bound in \Cref{lem:proper-finite-average}.

Throughout, use the parameters and functions from
\Cref{lem:proper-finite-average}.
Fix stored probes $\bz_1,\ldots,\bz_N$ satisfying
\[
 \norm{\bz_i}\le T_0
 \qquad\text{for every }i\le N,
\]
where $T_0\ge1$ is a supplied deterministic bound.
In the modified phase-1 construction,
\[
 T_0=\sqrt d+\sqrt{2\log(8A_0)},
\]
where $A_0$ is the confidence-independent coefficient in the
phase-1 sample bound.
All probabilities below concern only the fresh standard Gaussian
smoothing probes.

\subsection{Selected gradients and truncation error}

Recall that
\[
 v(y)=\min_{j\le N}\{u_{(j)}(y)+c_j\},
 \qquad
 u_i(y)=C_L(Q(\rho_s y+\sigma_s \bz_i)),
 \qquad c_j=\psi(j/N).
\]
The numbers $c_j$ are fixed constants.
Clipping, sorting, and taking a finite minimum preserve local
Lipschitzness. Moreover, $v$ selects one of the polynomial expressions
\[
 Q(\rho_s y+\sigma_s \bz_i)+c_j,\qquad -L+c_j,\qquad L+c_j.
\]
Resolve ties by a fixed ordering of the indices, and let
$d_v(y)$ be the gradient of the selected piece.
Nonidentical polynomial pieces agree only on a set of Lebesgue
measure zero, while identical pieces have identical gradients.
Consequently, $d_v=\nabla v$ almost everywhere.

By \Cref{lem:proper-gradient-growth}, uniformly for
$x\in\mathcal B_R$,
\begin{equation}\label{eq:finite-average-growth}
 \norm{d_v(ax+bz)}\le D(1+\norm z)^k,
 \qquad
 \left(\E_{\bz}\norm{d_v(ax+b\bz)}^2\right)^{1/2}\le D.
\end{equation}
The gradients are bounded by $D(1+\norm\bz)^k$, which has finite
expectation because $\bz$ is Gaussian.
This allows us to compute the gradients below by differentiating
inside the expectations.

Define the truncated expectation
\[
 \widetilde U(x)
 =\E_{\bz}\bigl[
   \one_{\{\norm\bz\le T\}}v(ax+b\bz)
 \bigr].
\]
Since $b>0$, the point $ax+b\bz$ has a density.
It therefore lies, almost surely, where $v$ is differentiable and
$d_v=\nabla v$. Thus
\[
 \nabla U(x)=a\,\E_{\bz}d_v(ax+b\bz),
 \qquad
 \nabla\widetilde U(x)
 =a\,\E_{\bz}\bigl[
   \one_{\{\norm\bz\le T\}}d_v(ax+b\bz)
 \bigr].
\]

\begin{lemma}[Truncation error]
\label{lem:finite-average-truncation}
Uniformly for $x\in\mathcal B_R$,
\begin{equation}\label{eq:finite-average-tail}
 |\widetilde U(x)-U(x)|\le\sqrt{\beta}/4,
 \qquad
 \norm{\nabla\widetilde U(x)-\nabla U(x)}
 \le\sqrt{\beta}/4.
\end{equation}
\end{lemma}

\begin{proof}
Gaussian concentration gives
\[
 p_{\mathrm{tail}}
 :=\Pr[\norm\bz>T]
 \le\exp\left(-\frac{(T-\sqrt d)^2}{2}\right)
 =\left(\frac{\sqrt{\beta}}{8(L+D)}\right)^2.
\]
Using $|v|\le L$ for values and
\eqref{eq:finite-average-growth} with Cauchy--Schwarz for gradients,
we obtain
\[
 \begin{aligned}
 |\widetilde U(x)-U(x)|
 &\le Lp_{\mathrm{tail}}\le\sqrt{\beta}/4,\\
 \norm{\nabla\widetilde U(x)-\nabla U(x)}
 &\le aD\sqrt{p_{\mathrm{tail}}}\le\sqrt{\beta}/4.
 \end{aligned}
\]
\end{proof}

\subsection{Pseudo-dimension of the sampled functions}

Consider the following functions of the probe $z$, indexed by
$x\in\mathcal B_R$:
\[
 F_x(z)=\one_{\{\norm z\le T\}}v(ax+bz),
 \qquad
 J_{x,\ell}(z)
 =\one_{\{\norm z\le T\}}a\,d_{v,\ell}(ax+bz),
 \quad 1\le\ell\le d,
\]
where $d_{v,\ell}$ is the $\ell$th coordinate of $d_v$.
They satisfy $|F_x|\le L$ and $|J_{x,\ell}|\le K$.
Their expectations are $\widetilde U(x)$ and
$\partial_\ell\widetilde U(x)$, respectively.

For independent probes $\bw_1,\ldots,\bw_M\sim\gauss$, define
\[
 A(x)=\frac1M\sum_{j=1}^M F_x(\bw_j)
 =\frac1M\sum_{j=1}^M
 \one_{\{\norm{\bw_j}\le T\}}v(ax+b\bw_j).
\]
The empirical averages of $F_x$ and $J_{x,\ell}$ are therefore
$A(x)$ and, for almost every $x$, $\partial_\ell A(x)$, respectively.

\begin{lemma}[Pseudo-dimension bound]
\label{lem:finite-average-pseudodimension}
Each of the value and coordinate derivative families has
pseudo-dimension at most
\[
 V=Cd\log((k+2)(N+1)).
\]
\end{lemma}

\begin{proof}
Write
\[
 P_i(x,z)=Q\bigl(\rho_s(ax+bz)+\sigma_s\bz_i\bigr).
\]
The signs of $P_i-L$, $P_i+L$, and $P_i-P_h$ determine clipping
and sorting, including the fixed rule at ties.
Every score considered in the subsequent minimum is one of
\[
 P_i(x,z)+c_j,\qquad -L+c_j,\qquad L+c_j.
\]
There are $O((N+1)^2)$ such expressions, so their pairwise
comparisons require $O((N+1)^4)$ distinct polynomial tests.

For the class $\{F_x:x\in\mathcal B_R\}$, consider the test
$F_x(z)>\tau$, which asks whether the value at probe $z$ exceeds
the threshold $\tau$.
Here $(z,\tau)$ is the input, and $x$ indexes the function being
tested.
To apply \Cref{fact:goldberg-jerrum}, we express this test using
polynomial comparisons.
First check whether the probe passes the cutoff by testing
$T^2-\norm z^2\ge0$.
Outside the cutoff, $F_x(z)=0$, so the test reduces to $0>\tau$.
Inside the cutoff, use the comparisons above to select the active
polynomial expression and compare its value with $\tau$.

For the derivative family, use the same cutoff and piece-selection
tests.
The possible values of $J_{x,\ell}(z)$ are
\[
 a\rho_s\,\partial_\ell Q
 \bigl(\rho_s(ax+bz)+\sigma_s\bz_i\bigr)
 \quad\text{and}\quad 0.
\]
Comparing these values with $\tau$ determines whether
$J_{x,\ell}(z)>\tau$.

In either case, the predicate is a Boolean formula using at most
$C(N+1)^4$ distinct polynomial tests, each of degree at most
$\max\{k,2\}$ in $(x,z,\tau)$.
Only the $d$ coordinates of $x$ are parameters; $(z,\tau)$ is
the input.
Therefore, \Cref{fact:goldberg-jerrum} gives
\[
 \operatorname{Pdim}\{F_x:x\in\mathcal B_R\},
 \quad
 \operatorname{Pdim}\{J_{x,\ell}:x\in\mathcal B_R\}
 \le Cd\log((k+2)(N+1))=V.
\]
These semialgebraic families satisfy the measurability conditions
of \Cref{fact:bounded-uniform-convergence}.
Indeed, the polynomial tests make the families jointly Borel
measurable.
The relevant suprema are universally measurable, being suprema
over a Borel subset of a Euclidean parameter space of jointly
Borel functions.
\end{proof}

\subsection{Bounding the sample size}

We now verify the claimed sample-size bound.

\begin{lemma}[Sample-size bound]
\label{lem:finite-average-sample-size}
There is an integer $M$ satisfying \eqref{eq:Mchoice} and
\[
 M\le
 \left(
 \frac{(d+1)(k+1)(L+1)(1+T_0)}{\beta}
 \right)^{C(k+1)}
 \left[
   \log(N+1)+\log\frac2\delta
 \right].
\]
\end{lemma}

\begin{proof}
Set
\[
 H=\frac{(d+1)(k+1)(L+1)(1+T_0)}{\beta}.
\]
The definitions imply, with universal constants,
\[
 D\le H^{C(k+1)},\qquad
 1+T\le H^C,\qquad
 K\le H^{C(k+1)}.
\]
Also,
\[
 V=Cd\log((k+2)(N+1))
 \le H^C\bigl(1+\log(N+1)\bigr),
\]
and
\[
 \log\frac{Cd(L+K+1)}{\sqrt\beta}
 \le C(k+1)\log H.
\]
Taking $M$ to be the ceiling of the right-hand side of
\eqref{eq:Mchoice}, substituting these estimates, and increasing
the universal constant $C$ gives
\[
 M\le H^{C(k+1)}
 \left[\log(N+1)+\log\frac2\delta\right],
\]
as claimed.
\end{proof}

\subsection{Completing the uniform approximation}

\begin{proof}[Proof of \Cref{lem:proper-finite-average}]
We first control sampling error.
Apply \Cref{fact:bounded-uniform-convergence} with accuracy
$\sqrt{\beta}/2$ for the value family and
$\sqrt{\beta}/(2\sqrt d)$ for each derivative family,
and with failure probability $\delta/(2(d+1))$ for each family.
The bounds $|F_x|\le L$ and $|J_{x,\ell}|\le K$, together with
\Cref{lem:finite-average-pseudodimension}, show that
\eqref{eq:Mchoice} suffices for all these applications.
A union bound therefore gives, with probability at least
$1-\delta/2$,
\[
 \sup_{x\in\mathcal B_R}
 |A(x)-\widetilde U(x)|\le\sqrt{\beta}/2,
 \qquad
 \norm{\nabla A(x)-\nabla\widetilde U(x)}
 \le\sqrt{\beta}/2
 \quad\text{for a.e. }x\in\mathcal B_R.
\]
Combining these bounds with \Cref{lem:finite-average-truncation}
proves \eqref{eq:finite-bounds}.

Each summand defining $A$ is locally Lipschitz, since its cutoff
is constant as $x$ varies.
On the event above,
\[
 \norm{\nabla A(x)}\le\Lip(U)+\sqrt{\beta}
 \quad\text{for a.e. }x\in\mathcal B_R.
\]
Since $A$ is locally Lipschitz and $\mathcal B_R$ is convex,
the almost-everywhere gradient bound gives
\[
 \Lip(A|_{\mathcal B_R})\le\Lip(U)+\sqrt{\beta}.
\]
To justify this, mollify $A$ on smaller balls and integrate along
line segments.
Passing to the limit gives the bound in the interior, and
continuity extends it to the boundary.
Finally, \Cref{lem:finite-average-sample-size} gives the claimed
bound on $M$.
\end{proof}
\section{Agnostic learning under a variance bound}
\label[appendix]{app:agnostic}\label[appendix]{sec:unbounded}

We first extend the bounded-label algorithm to agnostic absolute loss.
We then remove boundedness by centering and clipping labels with known
variance bound. We use the centering and absolute-distance clipping estimates
from \Cref{sec:realizable-clipping}.

\begin{theorem}[Bounded-label agnostic learning under absolute loss]
\label{thm:agnostic-bounded}
For $\bx\sim\gauss$, arbitrary labels $|\by|\le B$, $B\ge1$,
$0<\eps\le1$, and $0<\delta<1/2$, there is an algorithm returning a
finite globally $1$-Lipschitz formula $H:\R^d\to[-B,B]$ with
$\err_1(H)\le\OPT_1+\eps$ with probability at least $1-\delta$.
Its sample, arithmetic training and evaluation costs, and representation
size are $(d+1)^{\poly(B,1/\eps)}\left(\log\frac2\delta\right)^3$.
\end{theorem}
\begin{proof}
Use \Cref{prop:polynomial-learning-agnostic} with $\ell=1$,
$\eta=\eps/2$, and failure probability $\delta/2$. Choose its defect
parameter as required by \Cref{prop:finite-repair-interface} with
$\tau=\eps/2$, and apply that correction independently with failure
probability $\delta/2$. Clipping cannot increase absolute loss, so
\[
 \err_1(H)\le \err_1(C_B(p))+\norm{H-C_B(p)}_1
 \le\OPT_1+\eps/2+\eps/2.
\]
The correction depends only on the polynomial's squared defect; it does
not require realizability. A union bound and the resource bounds of the
two steps finish the proof.
\end{proof}

\begin{theorem}[Agnostic absolute loss with a known variance bound]
\label{thm:unbounded-absolute}
Suppose $\bx\sim\gauss$ and $\Var(\by)\le V^2$ for known $V\ge1$.
For $0<\eps\le1$ and $0<\delta<1/2$, there is an algorithm returning
a finite $1$-Lipschitz formula $H$ with $\err_1(H)\le\OPT_1+\eps$
with probability at least $1-\delta$. Its sample, arithmetic training and
evaluation costs, and representation size are
$(d+1)^{\poly(V,1/\eps)}\left(\log\frac2\delta\right)^3$.
The mean may be arbitrary and unknown. After centering, we use threshold
$T_1=18V+\sqrt{2\log(8/\eps)}$.
\end{theorem}

\subsection{Learning from clipped labels}

\begin{proof}[Proof of \Cref{thm:unbounded-absolute}]
Let $m$ be the preliminary sample median from
Lemma~\ref{lem:unbounded-centering}. Write $c=\E \by$. Fix a value of $m$ with $|m-c|\le2V$, and
write $\bz=\by-m$ and $R=3V$, so $\norm \bz_2\le R$. Subtracting $m$ from both
labels and predictors preserves the loss and the Lipschitz constraint,
so the centered problem has the same optimum. To distinguish the losses
before and after clipping, write
$\err_{1,A}(v)=\E|v(\bx)-A|$ for a label variable $A$.

The algorithm uses threshold $T=T_1$ from the theorem and sets
$\bw=C_T(\bz)$. On fresh samples of $(\bx,\bw)$, it applies
\Cref{thm:agnostic-bounded} with accuracy $\eta=\eps/4$ and failure probability
$\delta/2$, obtaining $\widehat h:\R^d\to[-T,T]$. It returns
$H=m+\widehat h$.

Condition on success of the bounded-label learner and fix its output.
For any $0<\zeta\le\eps/4$, choose a $1$-Lipschitz comparator $f$ for the
centered problem with $\err_{1,\bz}(f)\le\OPT_1+\zeta$.
Its clipped version $u=C_T(f)$ is still $1$-Lipschitz, so the learner's
guarantee gives
\begin{equation}\label{eq:unbounded-absolute-comparison}
 \err_{1,\bw}(\widehat h)
 \le\err_{1,\bw}(u)+\eta.
\end{equation}
We need to transfer this comparison from $\bw$ to $\bz$, and then bound the
cost of replacing $f$ by $u$.

To bound the cost of clipping $f$, observe that the zero predictor has
loss at most $R$. Hence $\err_{1,\bz}(f)\le R+\zeta$, and the triangle
inequality gives $\norm f_1\le\norm \bz_1+\err_{1,\bz}(f)\le3R$.

We first transfer the comparison to the original centered labels. For every $v\in[-T,T]$ and $z\in\R$,
\[
 |v-z|=|v-C_T(z)|+|z-C_T(z)|.
\]
Since both $\widehat{h}$ and $u$ take values in $[-T,T]$, the last term cancels
when we subtract their risks. Thus \Cref{eq:unbounded-absolute-comparison}
implies $\err_{1,\bz}(\widehat h)\le\err_{1,\bz}(u)+\eta$.

It remains to bound the error from clipping $f$. Set $r=\sqrt{2\log(8/\eps)}>1$. Since $T_1=6R+r$ and
$\norm f_1\le3R$, Lemma~\ref{lem:l1-clipping} gives
$\norm{u-f}_1\le2e^{-r^2/2}/r\le\eps/4$.
Therefore
\[
 \err_1(H)=\err_{1,\bz}(\widehat h)
 \le\err_{1,\bz}(f)+\norm{u-f}_1+\eta
 \le\OPT_1+\zeta+\eps/4+\eta<\OPT_1+\eps.
\]

\paragraph{Success probability and resources.}
Adding $m$ preserves Lipschitzness. The centering bound fails with
probability at most $\delta/2$. For each median satisfying that bound, the
learner uses fresh samples and fails with probability at most $\delta/2$.
Thus both loss and properness hold with probability at least $1-\delta$.
The threshold is polynomially bounded in $V,1/\eps$, so substituting
$B=T_1$ and accuracy $\eps/4$ into \Cref{thm:agnostic-bounded} gives the
claimed resource bounds. Evaluation uses no further randomness.
\end{proof}

\section{Extension to squared loss}
\label{app:squared-loss}

The correction stages control the change in predictions in Gaussian $L_1$.
For bounded labels, this also controls the change in squared loss. We use
squared loss in the polynomial learner and keep the same correction stages.

\begin{theorem}[Bounded-label agnostic learning under squared loss]
\label{thm:squared-bounded}
Under the assumptions of \Cref{thm:agnostic-bounded}, an algorithm returns
a finite globally $1$-Lipschitz formula $H:\R^d\to[-B,B]$ satisfying
$\err_2(H)\le\OPT_2+\eps$ with probability at least $1-\delta$, with
complexity $(d+1)^{\poly(B,1/\eps)}\left(\log\frac2\delta\right)^3$.
In the realizable case this gives $\err_2(H)\le\eps$.
\end{theorem}
\begin{proof}
For $a,b,y\in[-B,B]$,
\[
 |(a-y)^2-(b-y)^2|=|a-b|\,|a+b-2y|\le4B|a-b|.
\]
Use \Cref{prop:polynomial-learning-agnostic} with $\ell=2$,
$\eta=\eps/2$, and failure probability $\delta/2$, choosing its defect
parameter for \Cref{prop:finite-repair-interface} with
$\tau=\eps/(8B)$. Apply that correction with failure probability
$\delta/2$. Since clipping cannot increase squared loss against bounded
labels, on the joint success event
\[
 \err_2(H)\le \err_2(C_B(p))+4B\norm{H-C_B(p)}_1
 \le\OPT_2+\eps/2+4B\tau=\OPT_2+\eps.
\]
All accuracy parameters are polynomially related to $B,1/\eps$, so the
resource bounds follow as in the absolute-loss case.
\end{proof}

To handle unbounded labels under squared loss, we use a fourth-moment
bound for Lipschitz predictors. Centering is supplied by
\Cref{lem:unbounded-centering} in the main body.

\begin{lemma}[Fourth moments and clipping]
\label{lem:unbounded-fourth-moment}
Let $f:\R^d\to\R$ be $1$-Lipschitz. Then
$\norm f_4^4\le\norm f_2^4+4\norm f_2^2$.
In particular, if $\norm f_2\le3R$ for $R\ge1$, then $\norm f_4\le4R$ and
\[
 \norm{f-C_T(f)}_2\le\frac{16R^2}{T}\qquad(T>0).
\]
\end{lemma}
\begin{proof}
We apply the Gaussian Poincar\'e inequality to $f^2$. This is valid because
$f$ has at most linear growth, so $f^2$ and its weak gradient are square
integrable under Gaussian measure. We obtain
\[
 \E f^4-(\E f^2)^2
 =\Var(f^2)\le\E\norm{\nabla(f^2)}_2^2\le4\E f^2,
\]
where the last inequality uses the chain rule and Lipschitzness of $f$.
Thus $\norm f_4^4\le81R^4+36R^2\le256R^4$ when $\norm f_2\le3R$.
Finally, $(|z|-T)_+\le z^2/T$ implies
$\norm{f-C_T(f)}_2\le\norm f_4^2/T$.
\end{proof}

\begin{theorem}[Squared loss with a known variance bound]
\label{thm:unbounded-squared}
Under the assumptions of \Cref{thm:unbounded-absolute}, there is an
algorithm returning a finite globally $1$-Lipschitz formula $H$ with
$\err_2(H)\le\OPT_2+\eps$ with probability at least $1-\delta$.
Its sample, arithmetic training and evaluation costs, and representation
size are $(d+1)^{\poly(V,1/\eps)}\left(\log\frac2\delta\right)^3$.
After centering, we use $T_2=4096(3V)^5/\eps^2$.
\end{theorem}

A second-moment bound need not make the squared label-clipping error
small. We instead compare the risks of two Lipschitz predictors: the
label-square terms cancel, and the fourth-moment estimate above controls
the remaining product.

\begin{proof}[Proof of \Cref{thm:unbounded-squared}]
Let $m$ be the preliminary sample median from
Lemma~\ref{lem:unbounded-centering}. Write $c=\E \by$. Fix a value of $m$ with $|m-c|\le2V$, and
write $\bz=\by-m$ and $R=3V$, so $\norm \bz_2\le R$. Subtracting $m$ from both
labels and predictors preserves the loss and the Lipschitz constraint,
so the centered problem has the same optimum. To distinguish the losses
before and after clipping, write
$\err_{2,A}(v)=\E|v(\bx)-A|^2$ for a label variable $A$.

The algorithm uses threshold $T=T_2$ from the theorem and sets
$\bw=C_T(\bz)$. On fresh samples of $(\bx,\bw)$, it applies
\Cref{thm:squared-bounded} with accuracy $\eta=\eps/4$ and failure probability
$\delta/2$, obtaining $\widehat h:\R^d\to[-T,T]$. It returns
$H=m+\widehat h$.

Condition on success of the bounded-label learner and fix its output.
For any $0<\zeta\le\eps/4$, choose a $1$-Lipschitz comparator $f$ for the
centered problem with $\err_{2,\bz}(f)\le\OPT_2+\zeta$.
Its clipped version $u=C_T(f)$ is still $1$-Lipschitz, so the learner's
guarantee gives
\begin{equation}\label{eq:unbounded-squared-comparison}
 \err_{2,\bw}(\widehat h)
 \le\err_{2,\bw}(u)+\eta.
\end{equation}
We need to transfer this comparison from $\bw$ to $\bz$, and then bound the
cost of replacing $f$ by $u$.

To apply the clipping estimates, we first bound the sizes of the
predictors. The zero predictor has loss at most $R^2$ against either
$\bz$ or $\bw$. Thus
\[
 \err_{2,\bz}(f)\le R^2+\zeta,
 \qquad
 \err_{2,\bw}(\widehat h)\le R^2+\eta.
\]
The triangle inequality gives
$\norm f_2\le\norm \bz_2+
\err_{2,\bz}(f)^{1/2}\le3R$, and the same argument applies to
$\widehat h$ with labels $\bw$. Since clipping decreases the norm of $f$,
we have
\[
 \norm f_2,\quad\norm u_2,\quad\norm{\widehat h}_2\le3R.
\]

We compare the risks of $\widehat h$ and $u$ before and after clipping
the labels. The terms $\bz^2$ and $\bw^2$ cancel when we subtract the risks,
but the terms involving both labels and predictions leave a difference:
\begin{align*}
 \err_{2,\bz}(\widehat h)-\err_{2,\bz}(u)
 &=\err_{2,\bw}(\widehat h)-\err_{2,\bw}(u)\\
 &\quad-2\E[(\bz-\bw)(\widehat h(\bx)-u(\bx))].
\end{align*}
To use the learner's guarantee for $\bw$, we must bound this extra term.
The labels can be large, but $\bz-\bw$ vanishes unless $|\bz|>T$. We will use
the small probability of this event together with a fourth-moment bound
on the difference between the predictors.

Let $E=\{|\bz|>T\}$. Since $\E \bz^2\le R^2$, Markov's inequality gives
$\Pr(E)\le R^2/T^2$. Also, $|\bz-\bw|\le|\bz|\mathbf1_E$.
Both $\widehat h$ and $u$ are $1$-Lipschitz and have $L_2$ norm at most
$3R$, so Lemma~\ref{lem:unbounded-fourth-moment} bounds each of their
$L_4$ norms by $4R$. Thus $\norm{\widehat h-u}_4\le8R$.
Applying Cauchy--Schwarz first to separate the label from the predictions,
and then to restrict the squared prediction difference to $E$, gives
\begin{align*}
 \bigl|\E[(\bz-\bw)(\widehat h(\bx)-u(\bx))]\bigr|
 &\le R\bigl(\E[(\widehat h(\bx)-u(\bx))^2\mathbf1_E]\bigr)^{1/2}\\
 &\le R\bigl(\E|\widehat h(\bx)-u(\bx)|^4\bigr)^{1/4}\Pr(E)^{1/4}\\
 &\le\frac{8R^{5/2}}{\sqrt T}.
\end{align*}
This bound uses no independence between the input and the label.
Substituting it and \Cref{eq:unbounded-squared-comparison} into the risk
identity, and using $T=T_2=4096R^5/\eps^2$, we obtain
\[
 \err_{2,\bz}(\widehat h)
 \le\err_{2,\bz}(u)+\eta+\frac{16R^{5/2}}{\sqrt{T_2}}
 =\err_{2,\bz}(u)+\eta+\eps/4.
\]

It remains to compare the clipped comparator $u$ with $f$ on the original
centered labels. Lemma~\ref{lem:unbounded-fourth-moment} gives
$a:=\norm{u-f}_2\le16R^2/T$. Expanding the square in
$u-\bz=(f-\bz)+(u-f)$ and applying Cauchy--Schwarz yields
\begin{align*}
 \err_{2,\bz}(u)-\err_{2,\bz}(f)
 &=2\E[(f(\bx)-\bz)(u(\bx)-f(\bx))]+a^2\\
 &\le2\sqrt{\err_{2,\bz}(f)}\,a+a^2\\
 &\le4Ra+a^2
 \le\frac{64R^3}{T}+\frac{256R^4}{T^2}
 \le\eps/4.
\end{align*}
Here we used $\err_{2,\bz}(f)\le R^2+\zeta$ and the choice of $T_2$.
The two clipping steps therefore cost at most $\eps/4$ each. Combining
them with the learner's error $\eta$ and the comparator's error $\zeta$
gives
\[
 \err_2(H)=\err_{2,\bz}(\widehat h)
 \le\err_{2,\bz}(f)+\eta+\eps/2
 \le\OPT_2+\zeta+\eta+\eps/2
 \le\OPT_2+\eps.
\]

\paragraph{Success probability and resources.}
Adding $m$ preserves Lipschitzness. The centering bound fails with
probability at most $\delta/2$. For each median satisfying that bound, the
learner uses fresh samples and fails with probability at most $\delta/2$.
Thus both loss and properness hold with probability at least $1-\delta$.
The threshold is polynomially bounded in $V,1/\eps$, so substituting
$B=T_2$ and accuracy $\eps/4$ into \Cref{thm:squared-bounded} gives the
claimed resource bounds. Evaluation uses no further randomness.
\end{proof}

\section{Why a label scale assumption is necessary}\label{app:scale-obstructions}
The agnostic learning result assumes a known bound on the variance
of the labels. The one-dimensional examples below show that some known
control of label scale is necessary for a uniform sample bound. Each
example has bounded labels, but their bound grows with the proposed
sample size.

\subsection{Squared loss}

Fix a proposed sample bound $n$ and $0<\eps\le1$.  Choose a Gaussian
measurable set $A$ of mass $q\le1/(10n)$ and compare targets
\[
 f_0=0,
 \qquad f_1=M\1_A,
 \qquad qM=4\sqrt\eps.
\]
For every $1$-Lipschitz $h$, translation by a constant remains admissible and
reduces its squared risk by exactly $(\E h-\E f)^2$ when its mean is adjusted
to the target mean.  Hence
\[
 \err_{2,f}(h)-\OPT_{2,f}\ge(\E h-\E f)^2.
\]
An $\eps$-optimal output for $f_0$ must therefore have mean in
$[-\sqrt\eps,\sqrt\eps]$, whereas one for $f_1$ must have mean in
$[3\sqrt\eps,5\sqrt\eps]$.  Couple both experiments using the same inputs
and algorithmic coins.  With probability at least $1-nq\ge9/10$, no input
hits $A$ and the transcripts coincide.  The two successful-output sets are
disjoint, so the two failure probabilities sum to at least $9/10$; one is
larger than $1/3$.

\subsection{Absolute loss}

Choose $0<\eta\le1/(20n)$ and put
\[
 t_\pm=\Phi^{-1}(1/2\pm\eta),
 \qquad
 f_\pm(x)=A_*\1_{\{x\ge t_\pm\}},
 \qquad
 A_*=4/\eta.
\]
If $m=\E h$ and $\Lip(h)\le1$, Gaussian Poincar\'e gives
$\E|h-m|\le\sqrt{\Var(h)}\le1$.  Consequently
\[
 |\err_{1,f_\pm}(h)-\err_{1,f_\pm}(m)|\le1.
\]
The best constant is $0$ for $f_+$ and $A_*$ for $f_-$.  For
$m\in[0,A_*]$, the two constant excess risks are
$2\eta m$ and $2\eta(A_*-m)$, whose sum is $8$; outside this interval the
sum is larger.  If a Lipschitz $h$ were $\eps$-optimal for one target, its
constant mean predictor would have excess at most $1+\eps$ above the best
constant.  No single $h$ can therefore succeed for both targets, since the
two constant excesses would sum to at most $2+2\eps\le4$.

The targets differ only on Gaussian mass $2\eta$.  Coupling the two
experiments makes their transcripts identical with probability at least
$1-2n\eta\ge9/10$, again forcing a constant failure probability for one of
them.

\bibliographystyle{alpha}
\bibliography{refs}
\end{document}